\documentclass[letterpaper]{article} % DO NOT CHANGE THIS
\usepackage{aaai24}  % DO NOT CHANGE THIS
\usepackage{times}  % DO NOT CHANGE THIS
\usepackage{helvet}  % DO NOT CHANGE THIS
\usepackage{courier}  % DO NOT CHANGE THIS
\usepackage[hyphens]{url}  % DO NOT CHANGE THIS
\usepackage{graphicx} % DO NOT CHANGE THIS
\usepackage{natbib}  % DO NOT CHANGE THIS AND DO NOT ADD ANY OPTIONS TO IT
\usepackage{caption} % DO NOT CHANGE THIS AND DO NOT ADD ANY OPTIONS TO IT
\usepackage{algorithm}
\usepackage{newfloat}
\usepackage{listings}
\DeclareCaptionStyle{ruled}{labelfont=normalfont,labelsep=colon,strut=off} % DO NOT CHANGE THIS
\floatstyle{ruled}
\newfloat{listing}{tb}{lst}{}
\floatname{listing}{Listing}
\title{\method: Trainable Global Prototypes with Adaptive-Margin-Enhanced Contrastive Learning for Data and Model Heterogeneity in Federated Learning}
\author {
    Jianqing Zhang\textsuperscript{\rm 1},
    Yang Liu\textsuperscript{\rm 2,3}\equalcontrib,
    Yang Hua\textsuperscript{\rm 4},
    Jian Cao\textsuperscript{\rm 1}\equalcontrib
}
\affiliations {
    \textsuperscript{\rm 1}Shanghai Jiao Tong University, 
    \textsuperscript{\rm 2}Institute for AI Industry Research, Tsinghua University, \\
    \textsuperscript{\rm 3}Shanghai Artificial Intelligence Laboratory
    \textsuperscript{\rm 4}Queen's University Belfast \\
    \{tsingz, cao-jian\}@sjtu.edu.cn, liuy03@air.tsinghua.edu.cn, y.hua@qub.ac.uk
}

\usepackage{graphicx}
\usepackage{caption}
\usepackage{subfigure}
\usepackage{makecell}
\usepackage{booktabs}
\usepackage{bm}
\usepackage{xcolor}
\usepackage{float}
\usepackage{multirow}
\usepackage{longtable}
\usepackage{framed}
\usepackage{enumitem}
\usepackage{amsthm,amsmath,amssymb}
\usepackage{pythonhighlight}
\usepackage{appendix}
\usepackage[utf8]{inputenc}
\usepackage{bbding}
\usepackage{tabularx}
\usepackage{dsfont}

\DeclareMathAlphabet{\mathpzc}{OT1}{pzc}{m}{it}

\usepackage{mathtools}
\usepackage{multirow}
\usepackage[noend]{algpseudocode}
\usepackage{color,soul}

\makeatletter
\DeclareRobustCommand\onedot{\futurelet\@let@token\@onedot}
\def\@onedot{\ifx\@let@token.\else.\null\fi\xspace}
\def\eg{\emph{e.g}\onedot} 
\def\ie{\emph{i.e}\onedot}

\makeatother

\definecolor{blue_}{RGB}{76, 114, 176}
\definecolor{orange_}{RGB}{221, 132, 82}
\definecolor{upload}{RGB}{47, 85, 151}
\definecolor{download}{RGB}{241, 13, 208}
\definecolor{red_}{RGB}{255, 0, 0}
\definecolor{gray_}{RGB}{127, 127, 127}
\definecolor{green_}{RGB}{1, 128, 0}
\definecolor{sjtured_}{RGB}{192, 0, 0}
\definecolor{sjtugreen_}{RGB}{84, 130, 53}
\definecolor{yellow_}{RGB}{255, 192, 0}

\usepackage{pifont}

\usepackage[capitalize]{cleveref}
\crefname{section}{Sec.}{Secs.}
\Crefname{section}{Section}{Sections}
\Crefname{table}{Table}{Tables}
\crefname{table}{Tab.}{Tabs.}

\usepackage{xspace}
\newcommand{\method}{FedTGP\xspace}
\newcommand{\tp}{TGP\xspace}
\newcommand{\acl}{ACL\xspace}

\begin{document}

\maketitle

\begin{abstract}
Recently, Heterogeneous Federated Learning (HtFL) has attracted attention due to its ability to support heterogeneous models and data. To reduce 
the high communication cost of transmitting model parameters, a major challenge in HtFL,
prototype-based HtFL methods are proposed to solely share class representatives, a.k.a, prototypes, among heterogeneous clients while maintaining the privacy of clients' models. 
However, these prototypes are naively aggregated into global prototypes on the server using weighted averaging, resulting in suboptimal global knowledge which negatively impacts the performance of clients. 
To overcome this challenge, we introduce a novel HtFL approach called \textbf{\method}, which leverages our \textbf{Adaptive-margin-enhanced Contrastive Learning (\acl)} to learn \textbf{Trainable Global Prototypes (\tp)} on the server. By incorporating \acl, our approach enhances prototype separability while preserving semantic meaning. Extensive experiments with twelve heterogeneous models demonstrate that our \method surpasses state-of-the-art methods by up to \textbf{9.08\%} in accuracy while maintaining the communication and privacy advantages of prototype-based HtFL. Our code is available at \url{https://github.com/TsingZ0/FedTGP}. 
\end{abstract}

\section{Introduction}

With the rapid increase in the amount of data required to train large models today, concerns over data privacy also rise sharply~\cite{shin2023scaling, li2021survey}. To facilitate training  machine learning models while protecting data privacy, Federated Learning (FL) has emerged as a new distributed machine learning paradigm~\cite{kairouz2019advances, li2020federated}.
However, in practical scenarios, traditional FL methods such as FedAvg~\cite{mcmahan2017communication} experience performance degradation when faced with statistical heterogeneity~\cite{t2020personalized, li2022federated}. Subsequently, personalized FL methods emerged to address the challenge of statistical heterogeneity by learning personalized model parameters. Nevertheless, most of them still assume the model architectures on all the clients are the same and communicate client model updates to the server to train a shared global model~\cite{zhang2022fedala, zhang2023gpfl, zhang2023eliminating, collins2021exploiting, li2021ditto}. These methods not only bring formidable communication cost~\cite{zhuang2023foundation} but also expose clients' models, which further raise privacy and intellectual property (IP) concerns~\cite{li2021survey, zhang2018protecting, wang2023model}. 

To alleviate these problems, Heterogeneous FL (HtFL)~\cite{tan2022fedproto} has emerged as a novel FL paradigm that enables clients to possess diverse model architectures and heterogeneous data without sharing private model parameters. 
Instead, various types of global knowledge are shared among clients to reduce communication and improve model performance. For example, some FL methods adopt knowledge distillation (KD) techniques~\cite{hinton2015distilling} and communicate predicted logits on a public dataset~\cite{li2019fedmd, lin2020ensemble, liao2023draftfed, zhang2021parameterized} as global knowledge for aggregation at the server. However, these methods highly depend on the availability and quality of the global dataset~\cite{zhang2023towards}. 
Data-free KD-based approaches utilize additional auxiliary models as global knowledge~\cite{wu2022communication, zhang2022fine}, but the communication overhead for sharing the auxiliary models is still considerable. 
Alternatively, prototype-based HtFL methods~\cite{tan2022fedproto, tan2022federated} propose to share lightweight class representatives, a.k.a, \textit{prototypes}, as global knowledge, significantly reducing communication overhead. 

\begin{figure*}[t]
	\centering
	\subfigure[The prototype margins in FedProto using Cifar10.]{\includegraphics[width=0.49\linewidth]{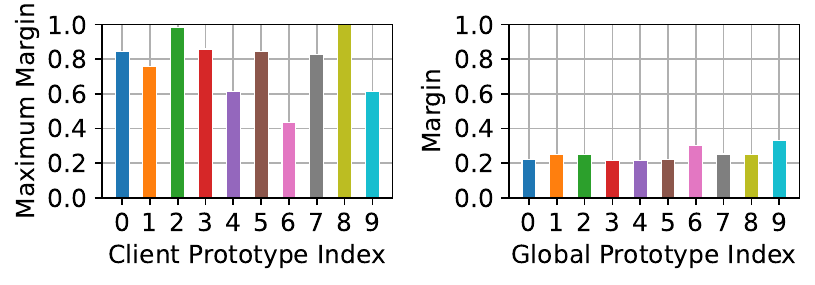}\label{fig:intro_fedproto}}
    \hfill
	\subfigure[The prototype margins in our \method using Cifar10.]{\includegraphics[width=0.49\linewidth]{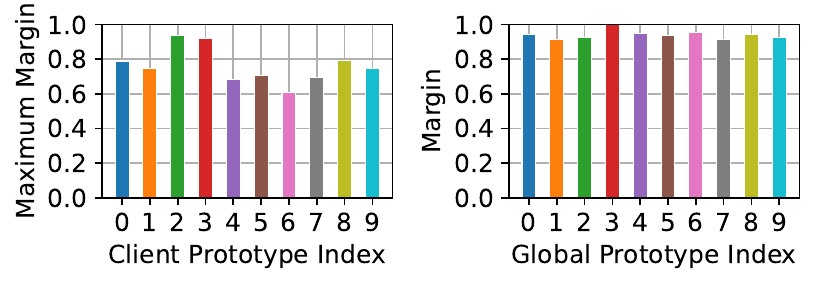}\label{fig:intro_our}}
	\caption{The illustration of the prototype margin change after generating global prototypes. The \textit{prototype margin} is the minimum Euclidean distance between the prototype of a specific class and the prototypes of other classes, and the maximum margin is the maximum prototype margin among all clients for each class. To enhance visualization and eliminate the influence of magnitude, we normalize the margin values for each method in these figures. Different colors represent different classes. (a) The global prototype margin \textit{shrinks} compared to the maximum of clients' prototype margins in FedProto. (b) The global prototype margin \textit{improves} compared to the maximum of clients' prototype margins in our \method.}
 \label{fig:intro}
\end{figure*}
However, existing prototype-based HtFL methods naively aggregate heterogeneous client prototypes on the server using weighted-averaging, which has several limitations. First, the weighted-averaging protocol requires clients to upload class distribution information of private data to the server as weights, which leaks sensitive distribution information about clients' data~\cite{yi2023fedgh}. 
Secondly, the prototypes generated from heterogeneous clients have diverse scales and separation margins. Averaging client prototypes generates uninformative global prototypes with smaller margins than the margins between well-separated prototypes. 
We demonstrate this ``\textit{prototype margin shrink}'' phenomenon in \cref{fig:intro_fedproto}. 
However, smaller margins between prototypes diminish their separability, ultimately generating poor prototypes~\cite{zhang2023semantic}. 

To address these limitations, we design a novel HtFL method using \textbf{Trainable Global Prototypes (\tp)}, termed \textbf{\method}, in which we train the desired global prototypes with our proposed \textbf{Adaptive-margin-enhanced Contrastive Learning (\acl)}. 
Specifically, we train the global prototypes to be separable while maintaining semantics via contrastive learning~\cite{hayat2019gaussian} with a specified margin. 
To avoid using an overlarge margin in early iterations and keep the best separability per iteration, we enhance contrastive learning by our adaptive margin, which reserves the maximum prototype margin among all clients in each iteration, as shown in \cref{fig:intro_our}. With the guidance of our separable global prototypes, \method can further enlarge the inter-class intervals for feature representations on each client. 

To evaluate the effectiveness of our \method, we conduct extensive experiments and compare it with six state-of-the-art methods in two popular statistically heterogeneous settings on four datasets using twelve heterogeneous models. Experimental results reveal that \method outperforms FedProto by up to \textbf{18.96\%} and surpasses other baseline methods by a large gap. Our contributions are:
\begin{itemize}
    \item We observe that naively averaging prototypes can result in ineffective global prototypes in FedProto-like schemes, as it causes the separation margin to shrink due to model heterogeneity in HtFL. 
    \item We propose an HtFL method called \method that learns trainable global prototypes with our adaptive-margin-enhanced contrastive learning technique to enhance inter-class separability. 
    \item Extensive comparison and ablation experiments on four datasets with twelve heterogeneous models demonstrate the superiority of \method over FedProto and other HtFL methods. 
\end{itemize}

\section{Related Work} 

\subsection{Heterogeneous Federated Learning} 

In recent times, Federated Learning (FL) has become a new machine learning paradigm that enables collaborative model training without exposing client data. Although personalized FL methods~\cite{t2020personalized, Zhang2023fedcp, yang2023dynamic, li2021ditto, collins2021exploiting} are proposed soon afterward to tackle the statistical heterogeneity of FL, they are still inapplicable for scenarios where clients own heterogeneous models for their specific tasks. Heterogeneous Federated Learning (HtFL) has emerged as a solution to support both model heterogeneity and statistical heterogeneity simultaneously, protecting both privacy and IP. 

One HtFL approach allows clients to sample diverse submodels from a shared global model architecture to accommodate the diverse communication and computing capabilities~\cite{diao2020heterofl, horvath2021fjord, wen2022federated}. 
However, concerns over sharing clients' model architectures still exist. Another HtFL approach is to split each client's model architecture and only share the top layers while allowing the bottom layers to have different architectures, \eg, LG-FedAvg~\cite{liang2020think} and FedGen~\cite{zhu2021data}. 
However, sharing and aggregating top layers may lead to unsatisfactory performance due to statistical heterogeneity~\cite{li2023no, luo2021no, wang2020federated}. Although learning a global generator can enhance the generalization ability ~\cite{zhu2021data}, its effectiveness highly relies on the quality of the generator. 

The above HtFL methods still require clients to have co-dependent model architectures. Alternatively, other methods seek to achieve HtFL with fully independent client models while communicating various kinds of information other than clients' models. Classic KD-based HtFL approaches~\cite{li2019fedmd, yu2022multimodal} share predicted knowledge on a global dataset to enable knowledge transfer among heterogeneous clients, but such a global dataset can be difficult to obtain~\cite{zhang2023towards}. 
%Among \textit{auxiliary-model-based methods}, 
FML~\cite{shen2020federated} and FedKD~\cite{wu2022communication} simultaneously train and share a small auxiliary model using mutual distillation~\cite{zhang2018deep} instead of using a global dataset. However, during the early iterations with poor feature-extracting abilities, the client model and the auxiliary model can potentially interfere with each other~\cite{li2023smkd}. 
Another popular approach is to share compact class representatives, \ie, prototypes. FedDistill~\cite{jeong2018communication} sends the class-wise logits from clients to the server and guides client model training by the globally averaged logits. 
FedProto~\cite{tan2022fedproto} and FedPCL~\cite{tan2022federated} share higher-dimensional prototypes instead of logits. However, all these approaches perform naive weighted-averaging on the clients' prototypes, resulting in subpar global prototypes due to statistical and model heterogeneity in HtFL. While FedPCL applies contrastive learning on each client for projection network training, it relies on pre-trained models, which is hard to satisfy in FL with private model architectures as clients join FL due to data scarcity~\cite{tan2022towards}. 
In this work, we explore methods to enhance the optimization of global prototypes, while maintaining the communication advantages inherent in such prototype-based approaches.  

\subsection{Trainable Prototype Learning} 
In centralized learning scenarios, trainable prototypes have been explored during model training to improve the intra-class compactness and inter-class discrimination of feature representations through the cross entropy loss~\cite{pinheiro2018unsupervised, yang2018robust} and regularizers~\cite{xu2020attribute, jin2010regularized}. 
Besides, some domain adaptation methods~\cite{tanwisuth2021prototype, kim2020attract} learn trainable global prototypes to transfer knowledge among domains. However, all these methods assume that data are nonprivate and the prototype learning depends on access to model and feature representations, which are infeasible in the FL setting. 

In our \method, we perform prototype learning on the server based solely on the knowledge of clients' prototypes, without accessing client models or features. 
In this way, the learning process of client models and global prototypes can be fully decoupled while mutually facilitating each other. 

\section{Method}

\subsection{Problem Statement and Motivation}

We have $M$ clients collaboratively train their models with heterogeneous architectures on their private and heterogeneous data $\{\mathcal{D}_i\}_{i=1}^M$. 
Following FedProto~\cite{tan2022fedproto}, we split each client $i$'s model into a feature extractor $f_i$ parameterized by $\theta_i$, which maps an input space $\mathbb{R}^D$ to a feature space $\mathbb{R}^K$, and a classifier $h_i$ parameterized by $w_i$, which maps the feature space to a class space $\mathbb{R}^C$. Clients collaborate by sharing global prototypes $\mathcal{P}$  with a server. Formally the overall collaborative training objective is
\begin{equation}
    \min_{\{\{\theta_i, w_i\}\}^M_{i=1}} \frac{1}{M} \sum^M_{i=1} \mathcal{L}_i(\mathcal{D}_i, \theta_i, w_i, \mathcal{P}). \label{eq:all}
\end{equation}

In FedProto, each client $i$ first obtains its prototype for each class $c$:
\begin{equation}
    P^c_i = \mathbb{E}_{({\bm x}, c) \sim \mathcal{D}_{i, c}} \ f_i({\bm x}; \theta_i), \label{eq:prototype}
\end{equation}
where $\mathcal{D}_{i, c}$ denotes the subset of $\mathcal{D}_i$ consisting of all data points belonging to class $c$. 
After receiving all prototypes from clients, the server then performs weighted-averaging for each class prototype: 
\begin{equation}
\bar{P}^c = \frac{1}{|\mathcal{N}_c|} \sum_{i \in \mathcal{N}_c} \frac{|\mathcal{D}_{i, c}|}{N_c} P^c_i, \label{eq:weighted} 
\end{equation}
where $\mathcal{N}_c$ and $N_c$ are the client set owning class $c$ and the total number of data of class $c$ among all clients. 
Next, the server transfers global information $\mathcal{P} = \{\bar{P}^c\}^C_{c=1}$ to each client, who performs guided training with a supervised loss
\begin{equation}
    \mathcal{L}_i := \mathbb{E}_{({\bm x}, y) \sim \mathcal{D}_i} \ell(h_i(f_i({\bm x} ;\theta_i); w_i), y) + \lambda \mathbb{E}_{c \sim \mathcal{C}_i} \phi(P^c_i, \bar{P}^c), \label{eq:proto_local}
\end{equation}
where $\ell$ is the loss for client tasks, $\lambda$ is a hyperparameter, and $\phi$ measures the Euclidean distance. 
$\mathcal{C}_i$ is a set of classes on the data of client $i$. Different clients may own different $\mathcal{C}$ in HtFL with heterogeneous data. 

\begin{figure}[t]
	\centering
	\subfigure[FedProto]{\includegraphics[width=0.44\linewidth]{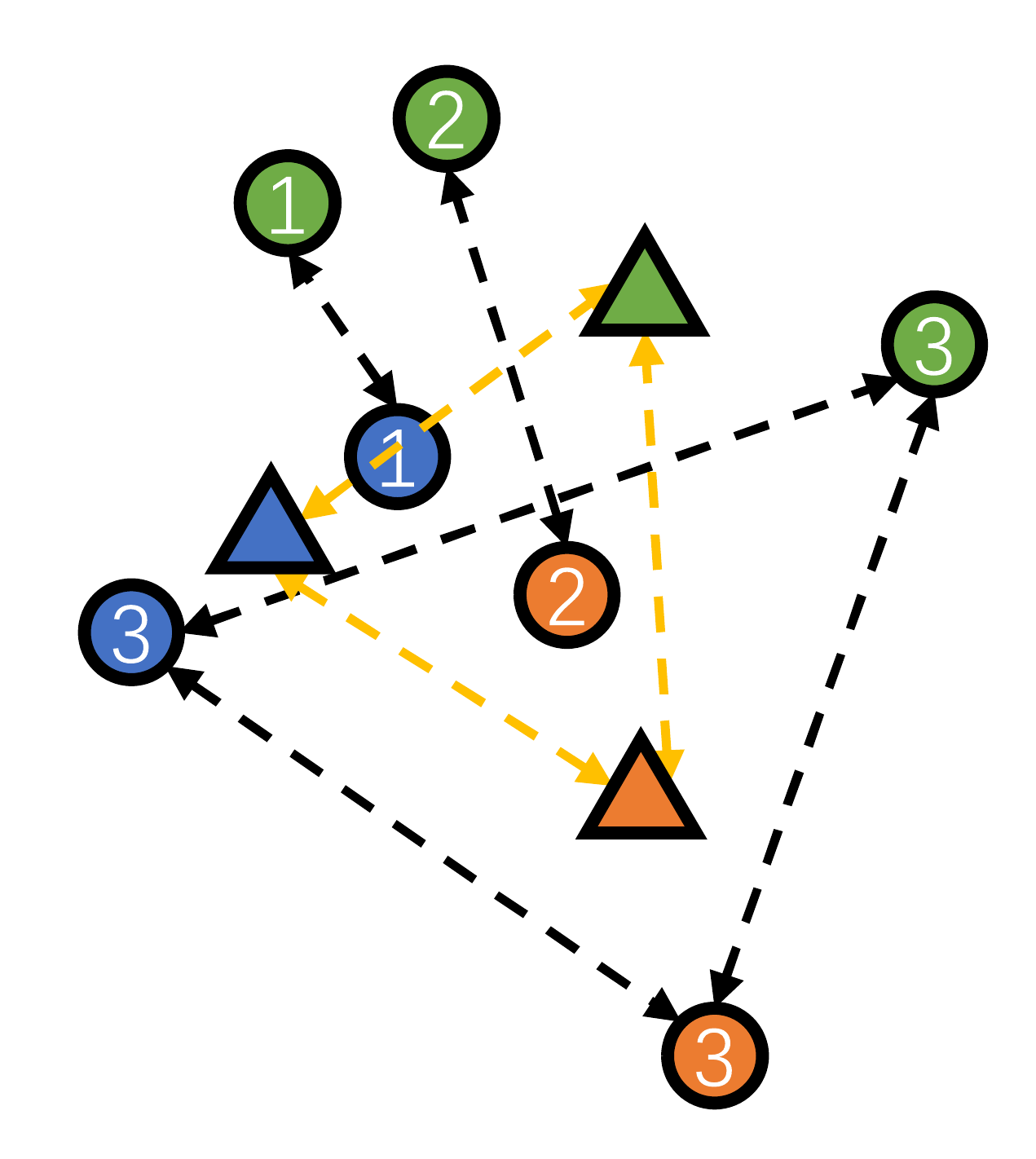}\label{fig:fedproto}}
    \hfill
	\subfigure[\method]{\includegraphics[width=0.53\linewidth]{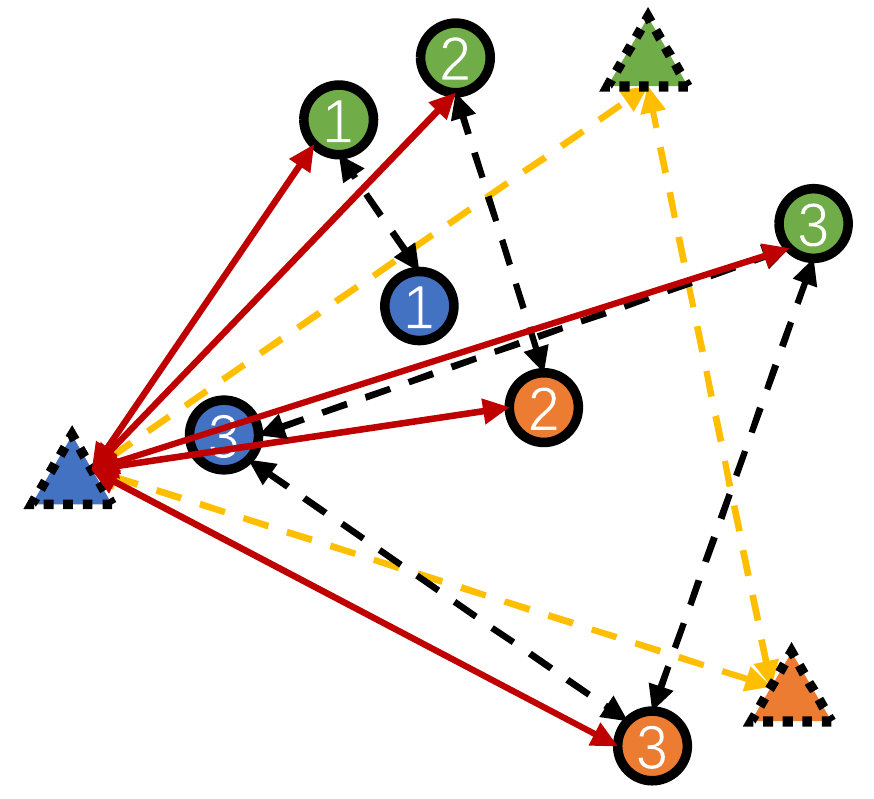}\label{fig:our}}
	\caption{The global and client prototypes in FedProto and our \method. Different colors and numbers represent classes and clients, respectively. 
 Circles represent the client prototypes and triangles represent the global prototypes. The black and \textcolor{yellow_}{yellow} dotted arrows show the inter-class separation among the client and global prototypes, respectively. 
 Triangles with dotted borders represent our \tp. 
 The \textcolor{sjtured_}{red} arrows show the inter-class intervals between \tp and the client prototypes of other classes in our \acl. } \label{fig:cmp}
\end{figure}

We observe that performing simple weighted-averaging to clients' prototypes in a heterogeneous environment may not generate desired information as expected, and we illustrate this phenomenon in \cref{fig:fedproto}. 
Due to the statistical and model heterogeneity, different clients extract much diverse feature representations of different classes with various separability and prototype margins. 
The weighted-averaging process assigns weights to client prototypes based solely on the amount of data, as indicated by \cref{eq:weighted}. However, since model performance in a heterogeneous environment can not be fully characterized by the data amount, prototypes generated by a poor client model may still be assigned a larger weight, causing the margin of global prototypes worse than the well-separated prototypes and impairing the training of the client models that previously produce well-separated prototypes. 

To address the above problem, we propose \method to (1) use Trainable Global Prototypes (\tp) with a separation objective on the server, (2) guide them to maintain large inter-class intervals with client prototypes while preserving semantics through our Adaptive-margin-enhanced Contrastive Learning (\acl) in each iteration, as shown in \cref{fig:our}, and (3) finally improve separability of different classes on each client with the guidance of separable global prototypes. 

% \subsection{Overview}

\subsection{Trainable Global Prototypes}

\begin{figure}[ht]
	\centering
    \includegraphics[width=\linewidth]{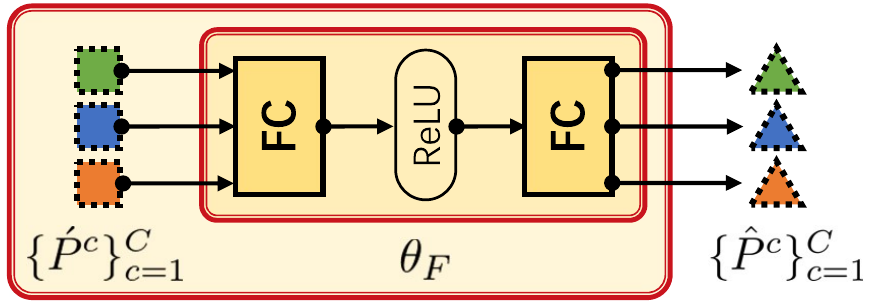}
	\caption{An example of trainable vectors ($\{\acute{P}^c\}^{C}_{c=1}$) and the further processing model ($\theta_{F}$). They only exist on the server.}
 \label{fig:F}
\end{figure}

In this section, we aim to learn a new set of global prototypes $\hat{\mathcal{P}} = \{\hat{P}^c\}^C_{c=1}$. Formally, we first randomly initialize a trainable vector $\acute{P}^c\in \mathbb{R}^K$ for each class $c$. Next, we place a neural network model $F$, parameterized by $\theta_{F}$, on the server to further process $\acute{P}^c$ to improve its training ability. 
The model $F$ transforms a given trainable vector to a global prototype with the same shape, \ie, $\forall c \in [C], \hat{P}^c = F(\acute{P}^c; \theta_{F})$, $\hat{P}^c \in \mathbb{R}^K$, as shown in \cref{fig:F}. 
$F$ consists of two Fully-Connected (FC) layers with a ReLU activation function in between. This structure is widely used for the server model in FL~\cite{chen2021bridging, shamsian2021personalized, ma2022layer}. 
In other words, the trainable global prototype $\hat{P}^c$ is parameterized by $\{\acute{P}^c, \theta_{F}\}$, and prototypes of different classes share the same parameter $\theta_{F}$.  

In order to learn effective prototypes, the trainable global prototype of class $c$ needs to achieve two goals: (1) closely align with the client prototypes of class $c$ to retain semantic information, and (2) maintain a significant distance from the client prototypes of other classes to enhance separability. The compactness and separation characteristics of contrastive learning~\cite{hayat2019gaussian, deng2019arcface} meet these two targets simultaneously. 
Thus, we can learn $\hat{\mathcal{P}}$ by 
\begin{equation}
     \min_{\hat{\mathcal{P}}} \ \sum^C_{c=1} \mathcal{L}^c_{P}, \label{eq:obj}
\end{equation}
\begin{equation}
    \mathcal{L}^c_{P} = \sum_{i\in \mathcal{I}^t} -\log \frac{e^{-\phi(P^c_i, \hat{P}^c)}}{e^{-\phi(P^c_i, \hat{P}^c)} + \sum_{c'} e^{-\phi(P^c_i, \hat{P}^{c'})}}, \label{eq:standard}
\end{equation}
where $c'\in [C], c'\ne c$, and $\mathcal{I}^t$ is the participating client set at $t$th iteration with client participation ratio $\rho$. 
Notice that all $C$ trainable global prototypes participate in the contrastive learning term in \cref{eq:standard}, which means they share pair-wise interactions with each other when performing gradient updates, and the gradient updates can be performed even with partial client participation. 

\subsection{Adaptive-Margin-Enhanced Contrastive Learning}

Although the standard contrastive loss \cref{eq:standard} can improve compactness and separation, it does not reduce intra-class variations. Moreover, the learned inter-class separation boundary may still lack clarity~\cite{choi2020amc}. 
To further improve the separability of global prototypes, we enforce a margin between classes when learning $\hat{\mathcal{P}}$. Inspired by the additive angular margin of ArcFace~\cite{deng2019arcface} used in an angular space for face recognition, we introduce a scalar $\delta$ to \cref{eq:standard} in our considered Euclidean space and rewrite $\mathcal{L}^c_{P}$ as
\begin{equation}
    \mathcal{L}^c_{P} = \sum_{i\in \mathcal{I}^t} -\log \frac{e^{-(\phi(P^c_i, \hat{P}^c) + \delta)}}{e^{-(\phi(P^c_i, \hat{P}^c) + \delta)} + \sum_{c'} e^{-\phi(P^c_i, \hat{P}^{c'})}}, \label{eq:margin}
\end{equation}
where $\delta > 0$. According to ~\cite{schroff2015facenet, hayat2019gaussian}, minimizing $\mathcal{L}^c_{P}$ is equivalent to minimizing $\tilde{\mathcal{L}}^c_{P}$, 
\begin{equation}
    \mathcal{L}^c_{P} \propto \tilde{\mathcal{L}}^c_{P} := \sum_{i\in \mathcal{I}^t} \sum_{c'} e^{\phi(P^c_i, \hat{P}^c) - \phi(P^c_i, \hat{P}^{c'}) + \delta}, 
\end{equation}
which reduces the distance between $P^c_i$ and $\hat{P}^c$ while increasing the distance between $P^c_i$ and $\hat{P}^{c'}$ with a margin $\delta$.

\begin{algorithm}[t]
    \caption{The learning process of \method.}
    \begin{algorithmic}[1]
        \Require 
        $M$ clients with their heterogeneous models and data, 
        trainable global prototypes $\hat{\mathcal{P}}$ on the server, 
        $\eta$: learning rate, 
        $T$: total communication iterations. 
        \Ensure 
        Well-trained client models. 
        \For{iteration $t=1, \ldots, T$}
            \State Server randomly samples a client subset $\mathcal{I}^t$. 
            \State Server sends $\hat{\mathcal{P}}$ to $\mathcal{I}^t$. 
            \For{Client $i \in \mathcal{I}^t$ in parallel}
                \State Client $i$ updates its model with \cref{eq:training}. 
                \State Client $i$ calculates prototypes $\mathcal{P}_i$ by \cref{eq:prototype}.
                \State Client $i$ sends $\mathcal{P}_i$ to the server. 
            \EndFor
            \State Server obtains $\delta(t)$ through \cref{eq:delta}
            \State Server updates $\hat{\mathcal{P}}$ with \cref{eq:deltat}. 
        \EndFor
        \\
        \Return Client models.
    \end{algorithmic}
    \label{algo}
\end{algorithm}

However, we observe that setting a large $\delta$ in early iterations may also mislead both the prototype training and the client model training because the feature extraction abilities of heterogeneous models are poor in the beginning. To retain the best separability of client prototypes within the semantic region in each iteration, we set the adaptive $\delta(t)$ to be the maximum margin among client prototypes of different classes with a threshold $\tau$,
\begin{equation}
    \delta(t) = \min(\max_{c \in [C], c' \in [C], c \ne c'} \phi(Q^c_t, Q^{c'}_t), \tau), \label{eq:delta}
\end{equation}
where $Q^c_t = \frac{1}{|\mathcal{P}^c_t|} \sum_{i\in \mathcal{I}^t} P^c_i, \forall c \in [C]$ represents the cluster center of the client prototypes for each class, and it differs from the weighted average $\bar{P}^c$ which adopts private distribution information as weights. $\mathcal{P}^c_t = \{P^c_i\}_{i\in\mathcal{I}^t}$, and $\tau$ is used to keep the margin from growing to infinite. 
Thus, we have
\begin{equation}
    \mathcal{L}^c_{P} = \sum_{i\in \mathcal{I}^t}-\log \frac{e^{-(\phi(P^c_i, \hat{P}^c) + \delta(t))}}{e^{-(\phi(P^c_i, \hat{P}^c) + \delta(t))} + \sum_{c'} e^{-\phi(P^c_i, \hat{P}^{c'})}}. \label{eq:deltat}
\end{equation}

\begin{table*}[ht]
  \caption{The test accuracy (\%) on four datasets in the pathological and practical settings using the HtFE$_8$ model group.}
  \centering
  \resizebox{\linewidth}{!}{
    \begin{tabular}{l|*{4}{c}|*{4}{c}}
    \toprule
    Settings & \multicolumn{4}{c|}{Pathological Setting} & \multicolumn{4}{c}{Practical Setting} \\
    \midrule
    Datasets & Cifar10 & Cifar100 & Flowers102 & Tiny-ImageNet & Cifar10 & Cifar100 & Flowers102 & Tiny-ImageNet \\
    \midrule
    LG-FedAvg & 86.82$\pm$0.26 & 57.01$\pm$0.66 & 58.88$\pm$0.28 & 32.04$\pm$0.17 & 84.55$\pm$0.51 & 40.65$\pm$0.07 & 45.93$\pm$0.48 & 24.06$\pm$0.10 \\
    FedGen & 82.83$\pm$0.65 & 58.26$\pm$0.36 & 59.90$\pm$0.15 & 29.80$\pm$1.11 & 82.55$\pm$0.49 & 38.73$\pm$0.14 & 45.30$\pm$0.17 & 19.60$\pm$0.08 \\
    FML & 87.06$\pm$0.24 & 55.15$\pm$0.14 & 57.79$\pm$0.31 & 31.38$\pm$0.15 & 85.88$\pm$0.08 & 39.86$\pm$0.25 & 46.08$\pm$0.53 & 24.25$\pm$0.14 \\
    FedKD & 87.32$\pm$0.31 & 56.56$\pm$0.27 & 54.82$\pm$0.35 & 32.64$\pm$0.36 & 86.45$\pm$0.10 & 40.56$\pm$0.31 & 48.52$\pm$0.28 & 25.51$\pm$0.35 \\
    FedDistill & 87.24$\pm$0.06 & 56.99$\pm$0.27 & 58.51$\pm$0.34 & 31.49$\pm$0.38 & 86.01$\pm$0.31 & 41.54$\pm$0.08 & 49.13$\pm$0.85 & 24.87$\pm$0.31 \\
    FedProto & 83.39$\pm$0.15 & 53.59$\pm$0.29 & 55.13$\pm$0.17 & 29.28$\pm$0.36 & 82.07$\pm$1.64 & 36.34$\pm$0.28 & 41.21$\pm$0.22 & 19.01$\pm$0.10 \\
    \midrule
    \method & \textbf{90.02$\pm$0.30} & \textbf{61.86$\pm$0.30} & \textbf{68.98$\pm$0.43} & \textbf{34.56$\pm$0.27} & \textbf{88.15$\pm$0.43} & \textbf{46.94$\pm$0.12} & \textbf{53.68$\pm$0.31} & \textbf{27.37$\pm$0.12} \\
    \bottomrule
    \end{tabular}}
    \label{tab:datasets_acc}
\end{table*}

\subsection{\method Framework}

We show the entire learning process of our \method framework in \cref{algo}. 
With the well-trained separable global prototypes, we send them to participating clients in the next iteration and guide client training with them to improve separability locally among feature representations of different classes by minimizing the client loss $\mathcal{L}_i$ for client $i$, 
\begin{equation}
    \mathcal{L}_i := \mathbb{E}_{({\bm x}, y) \sim \mathcal{D}_i} \ell(h_i(f_i({\bm x} ;\theta_i); w_i), y) + \lambda \mathbb{E}_{c \sim \mathcal{C}_i} \phi(P^c_i, \hat{P}^c), \label{eq:training}
\end{equation}
which is similar to \cref{eq:proto_local} but using the well-trained separable global prototypes $\hat{P}^c$ instead of $\bar{P}^c$. 
Following FedProto, we also utilize the global prototypes for inference on clients. Specifically, for a given input on one client, we calculate the $\phi$ distance between the feature representation and $C$ global prototypes, and then this input belongs to the class of the closest global prototype. 

Since our \method follows the same communication protocol as FedProto by transmitting only compact 1D-class prototypes, it naturally brings benefits to both privacy preservation and communication efficiency. 
Specifically, no model parameter is shared and the generation of low-dimensional prototypes is irreversible, preventing data leakage from inversion attacks. In addition, our \method does not require clients to upload the private class distribution information (\ie, $|\mathcal{D}_{i, c}|$ in \cref{eq:weighted}) to the server anymore, leading to less information revealed than FedProto. 

\section{Experiments}

\begin{table*}[t]
  \caption{The test accuracy (\%) on Cifar100 in the practical setting using heterogeneous feature extractors, heterogeneous classifiers, or a large number of clients ($\rho=0.5$) with the HtFE$_8$ model group. ``Res'' is short for ResNet. }
  \centering
  \resizebox{\linewidth}{!}{
    \begin{tabular}{l|*{4}{c}|*{2}{c}|*{2}{c}}
    \toprule
    Settings & \multicolumn{4}{c|}{Heterogeneous Feature Extractors} & \multicolumn{2}{c|}{Heterogeneous Classifiers} & \multicolumn{2}{c}{Large Client Amount}\\
    \midrule
    & HtFE$_2$ & HtFE$_3$ & HtFE$_4$ & HtFE$_9$ & Res34-HtC$_4$ & HtFE$_8$-HtC$_4$ & 50 Clients & 100 Clients \\
    \midrule
    LG-FedAvg & 46.61$\pm$0.24 & 45.56$\pm$0.37 & 43.91$\pm$0.16 & 42.04$\pm$0.26 & --- & --- & 37.81$\pm$0.12 & 35.14$\pm$0.47 \\
    FedGen & 43.92$\pm$0.11 & 43.65$\pm$0.43 & 40.47$\pm$1.09 & 40.28$\pm$0.54 & --- & --- & 37.95$\pm$0.25 & 34.52$\pm$0.31 \\
    FML & 45.94$\pm$0.16 & 43.05$\pm$0.06 & 43.00$\pm$0.08 & 42.41$\pm$0.28 & 41.03$\pm$0.20 & 39.23$\pm$0.42 & 38.47$\pm$0.14 & 36.09$\pm$0.28 \\
    FedKD & 46.33$\pm$0.24 & 43.16$\pm$0.49 & 43.21$\pm$0.37 & 42.15$\pm$0.36 & 39.77$\pm$0.42 & 40.59$\pm$0.51 & 38.25$\pm$0.41 & 35.62$\pm$0.55 \\
    FedDistill & 46.88$\pm$0.13 & 43.53$\pm$0.21 & 43.56$\pm$0.14 & 42.09$\pm$0.20 & 44.72$\pm$0.13 & 41.67$\pm$0.06 & 38.51$\pm$0.36 & 36.06$\pm$0.24 \\
    FedProto & 43.97$\pm$0.18 & 38.14$\pm$0.64 & 34.67$\pm$0.55 & 32.74$\pm$0.82 & 32.26$\pm$0.18 & 25.57$\pm$0.72 & 33.03$\pm$0.42 & 28.95$\pm$0.51 \\
    \midrule
    \method & \textbf{49.82$\pm$0.29} & \textbf{49.65$\pm$0.37} & \textbf{46.54$\pm$0.14} & \textbf{48.05$\pm$0.19} & \textbf{48.18$\pm$0.27} & \textbf{44.53$\pm$0.16} & \textbf{43.17$\pm$0.23} & \textbf{41.57$\pm$0.30} \\
    \bottomrule
    \end{tabular}}
    \label{tab:hetero}
\end{table*}

\subsection{Setup}

\noindent\textbf{Datasets. \ } We evaluate four popular image datasets for the multi-class classification tasks, including Cifar10 and Cifar100~\cite{krizhevsky2009learning}, Tiny-ImageNet~\cite{chrabaszcz2017downsampled} (100K images with 200 classes), and Flowers102~\cite{nilsback2008automated} (8K images with 102 classes). 

\noindent\textbf{Baseline methods. \ } To evaluate our proposed \method, we compare it with six popular methods that are applicable in HtFL, including LG-FedAvg~\cite{liang2020think}, FedGen~\cite{zhu2021data}, FML~\cite{shen2020federated}, FedKD~\cite{wu2022communication}, FedDistill~\cite{jeong2018communication}, and FedProto~\cite{tan2022fedproto}. 

\noindent\textbf{Model heterogeneity. \ } 
Unless explicitly specified, we evaluate the model heterogeneity regarding Heterogeneous Feature Extractors (HtFE). We use ``HtFE$_X$'' to denote the HtFE setting, where $X$ is the number of different model architectures in HtFL. We assign the $(i \mod X)$th model architecture to client $i$. %where $\mod$ is the modulo operation. 
For our main experiments, we use the ``HtFE$_8$'' model group with eight architectures including the 4-layer CNN~\cite{mcmahan2017communication}, GoogleNet~\cite{szegedy2015going}, MobileNet\_v2~\cite{sandler2018mobilenetv2}, ResNet18, ResNet34, ResNet50, ResNet101, and ResNet152~\cite{he2016deep}. 
To generate feature representations with an identical feature dimension $K$, we add an average pooling layer~\cite{szegedy2015going} after each feature extractor. By default, we set $K=512$. 

\noindent\textbf{Statistical heterogeneity. \ }  We conduct extensive experiments with two widely used statistically heterogeneous settings, the pathological setting~\cite{mcmahan2017communication, tan2022fedproto} and the practical setting~\cite{tan2022federated, li2021model, zhu2021data}. For the pathological setting, following FedAvg~\cite{mcmahan2017communication}, we distribute non-redundant and unbalanced data of 2/10/10/20 classes to each client from a total of 10/100/102/200 classes on Cifar10/Cifar100/Flowers102/Tiny-ImageNet datasets. For the practical setting, following MOON~\cite{li2021model}, we first sample $q_{c, i} \sim Dir(\beta)$ for class $c$ and client $i$, then we assign $q_{c, i}$ proportion of data points from class $c$ in a given dataset to client $i$, where $Dir(\beta)$ is the Dirichlet distribution and $\beta$ is set to 0.1 by default~\cite{lin2020ensemble}. 

\noindent\textbf{Implementation Details. \ } Unless explicitly specified, we use the following settings. We simulate a federation with 20 clients and a client participation ratio $\rho=1$. Following FedAvg, we run one training epoch on each client in each iteration with a batch size of 10 and a learning rate $\eta=0.01$ for 1000 communication iterations. We split the private data into a training set (75\%) and a test set (25\%) on each client. We average the results on clients' test sets and choose the best averaged result among iterations in each trial. For all the experiments, we run three trials and report the mean and standard deviation. We set $\lambda=0.1$ (the same as FedProto), $\tau=100$, and $S=100$ (the number of server training epochs) for our \method on all tasks. 
Please refer to the Appendix for more results and details.

\subsection{Performance}

As shown in \cref{tab:datasets_acc}, \method outperforms all the baselines on four datasets by up to \textbf{9.08\%} in accuracy. Specifically, using our \tp with \acl on the server, our \method can improve FedProto by up to \textbf{13.85\%}. The improvement is attributed to the enhanced separability of global prototypes. 
Besides, \method shows better performance in relatively harder tasks with more classes, as more classes mean more client prototypes, which benefits our global prototype training. 
However, the generator in FedGen does not consistently yield improvements in HtFL, as FedGen cannot outperform LG-FedAvg in all cases in \cref{tab:datasets_acc}.

\subsection{Impact of Model Heterogeneity}

To examine the impact of model heterogeneity in HtFL, we assess the performance of \method on four additional model groups with increasing model heterogeneity without changing the data distribution on clients: ``HtFE$_2$'' including the 4-layer CNN and ResNet18; ``HtFE$_3$'' including ResNet10~\cite{zhong2017deep}, ResNet18, and ResNet34; ``HtFE$_4$'' including the 4-layer CNN, GoogleNet, MobileNet\_v2, and ResNet18; ``HtFE$_9$'' including ResNet4, ResNet6, and ResNet8~\cite{zhong2017deep}, ResNet10, ResNet18, ResNet34, ResNet50, ResNet101, and ResNet152. We show results in \cref{tab:hetero}.

Our \method consistently outperforms other FL methods across various model heterogeneities by up to \textbf{5.64\%}, irrespective of the models’ sizes. We observe that all methods perform worse with larger model heterogeneity in HtFL. 
However, our \method only drops 1.77\%, while the decrease for the counterparts is 3.53\%$\sim$15.04\%, showing that our proposed \tp with \acl is more robust and less impacted by model heterogeneity. 

We further evaluate the scenarios with four Heterogeneous Classifiers (HtC$_4$)\footnote{Please refer to the Appendix for the details.} and create another two model groups: ``Res34-HtC$_4$'' uses the ResNet34 to build homogeneous feature extractors while both the feature extractors and classifiers are heterogeneous in ``HtFE$_8$-HtC$_4$''. 
We allocate classifiers to clients using the method introduced in HtFE$_X$. 
Since LG-FedAvg and FedGen require using homogeneous classifiers, these methods are not applicable here. In \cref{tab:hetero}, our \method still keeps the superiority in these scenarios. In the most heterogeneous scenario HtFE$_8$-HtC$_4$, our \method surpasses FedProto by \textbf{18.96\%} in accuracy with our proposed \tp and \acl. 

\subsection{Partial Participation with More Clients}

Additionally, we evaluate our method on the Cifar100 dataset with 50 and 100 clients, respectively, using partial client participation. When assigning Cifar100 to more clients using HtFE$_8$, the data amount on each client decreases, so all the methods perform worse with a larger client amount. Besides, we only sample half of the clients to participate in training in each iteration, \ie, $\rho=0.5$. 
In \cref{tab:hetero}, the superiority of our \method is more obvious with more clients. Specifically, our \method outperforms other methods by 4.66\% and 5.48\% with 50 clients and 100 clients, respectively. 

\subsection{Impact of Number of Client Training Epochs}

\begin{table}[ht]
  \caption{The test accuracy (\%) on Cifar100 in the practical setting using the HtFE$_8$ model group with a different number of client training epochs ($E$).}
  \centering
  \resizebox{\linewidth}{!}{
    \begin{tabular}{l|*{3}{c}}
    \toprule
     & $E=5$ & $E=10$ & $E=20$ \\
    \midrule
    LG-FedAvg & 40.33$\pm$0.15 & 40.46$\pm$0.08 & 40.93$\pm$0.23 \\
    FedGen & 40.00$\pm$0.41 & 39.66$\pm$0.31 & 40.07$\pm$0.12 \\
    FML & 39.08$\pm$0.27 & 37.97$\pm$0.19 & 36.02$\pm$0.22 \\
    FedKD & 41.06$\pm$0.13 & 40.36$\pm$0.20 & 39.08$\pm$0.33 \\
    FedDistill & 41.02$\pm$0.30 & 41.29$\pm$0.23 & 41.13$\pm$0.41 \\
    FedProto & 38.04$\pm$0.52 & 38.13$\pm$0.42 & 38.74$\pm$0.51 \\
    \midrule
    \method & \textbf{46.44$\pm$0.26} & \textbf{46.59$\pm$0.31} & \textbf{46.65$\pm$0.29} \\
    \bottomrule
    \end{tabular}}
    \label{tab:largeE}
\end{table}

During collaborative learning in FL, clients can alleviate the communication burden by conducting more client model training epochs before transmitting their updated models to the server~\cite{mcmahan2017communication}. However, we notice that increasing the number of client training epochs leads to reduced accuracy in methods such as FML and FedKD, which employ an auxiliary model. This decrease in accuracy can be attributed to the increased heterogeneity in the parameters of the shared auxiliary model before server aggregation. 
In contrast, other methods such as our proposed \method, can maintain their performance with more client training epochs.

\subsection{Impact of Feature Dimensions}

\begin{table}[ht]
  \caption{The test accuracy (\%) on Cifar100 in the practical setting using the HtFE$_8$ model group with different feature dimensions ($K$).}
  \centering
  \resizebox{\linewidth}{!}{
    \begin{tabular}{l|*{3}{c}}
    \toprule
     & $K=64$ & $K=256$ & $K=1024$ \\
    \midrule
    LG-FedAvg & 39.69$\pm$0.25 & 40.21$\pm$0.11 & 40.46$\pm$0.01 \\
    FedGen & 39.78$\pm$0.36 & 40.38$\pm$0.36 & 40.83$\pm$0.25 \\
    FML & 39.89$\pm$0.34 & 40.95$\pm$0.09 & 40.26$\pm$0.16 \\
    FedKD & 41.06$\pm$0.18 & 41.14$\pm$0.35 & 40.72$\pm$0.25 \\
    FedDistill & 41.69$\pm$0.10 & 41.66$\pm$0.15 & 40.09$\pm$0.27 \\
    FedProto & 30.71$\pm$0.65 & 37.16$\pm$0.42 & 31.21$\pm$0.27 \\
    \midrule
    \method & \textbf{46.28$\pm$0.59} & \textbf{46.30$\pm$0.39} & \textbf{45.98$\pm$0.38} \\
    \bottomrule
    \end{tabular}}
    \label{tab:K}
\end{table}

We also vary the feature dimension $K$ to evaluate its impact on model performance, as shown in \cref{tab:K}.  
We find that most methods show better performance with increasing feature dimensions from $K=64$ to $K=256$, but the performance degrades with an excessively large feature dimension, such as $K=1024$, as it becomes more challenging to train classifiers with too large feature dimension. In \cref{tab:K}, our \method achieves competitive performance with $K=64$, while FedProto lags by 6.45\% compared to $K=256$. 

\subsection{Communication Cost} 

\begin{table}[ht]
  \caption{The communication cost per iteration using the HtFE$_8$ model group on Cifar100 in the practical setting. 
  $\Theta$ represents the parameters for the auxiliary generator in FedGen. $\theta_g$ and $w_g$ denote the parameters of the auxiliary feature extractor and classifier, respectively, in FML and FedKD. $r$ is a compression rate introduced by SVD for parameter factorization in FedKD. $|\theta_g| \gg K\times C$. $C_i$ denotes the number of classes on client $i$. ``M'' is short for million. }
  \setlength{\tabcolsep}{3pt}
  \centering
  \resizebox{!}{!}{
    \begin{tabular}{l|cr}
    \toprule
     & Theory & Practice \\
    \midrule
    LG-FedAvg & $\sum^M_{i=1} |w_i|\times 2$ & 2.05M \\
    FedGen & $\sum^M_{i=1} (|w_i|\times 2 + |\Theta|)$ & 8.69M \\
    FML & $M\times (|\theta_g| + |w_g|)\times 2$ & 36.99M \\
    FedKD & $M\times (|\theta_g| + |w_g|)\times 2\times r$ & 33.04M \\
    FedDistill & $\sum^M_{i=1} C\times (C_i + C)$ & 0.29M \\
    FedProto & $\sum^M_{i=1} K\times (C_i + C)$ & 1.48M \\
    \midrule
    \method & $\sum^M_{i=1} K\times (C_i + C)$ & 1.48M \\
    \bottomrule
    \end{tabular}}
    \label{tab:comm}
\end{table}

We show the communication cost in \cref{tab:comm}. Specifically, we calculate the communication cost in both theory and practice. 
In \cref{tab:comm} FML and FedKD cost the most overhead in communication as they additionally transmit an auxiliary model. 
Although FedKD reduces the communication overhead through singular value decomposition (SVD) on the auxiliary model parameters, its communication cost is still much larger than prototype-based methods. 
In FedGen, downloading the generator from the server brings noticeable communication overhead. Although FedDistill costs $5.12\times$ less communication overhead than our \method, the information capacity of the logits is $5.12\times$ less than the prototypes, so FedDistill achieves lower accuracy than \method. 
In summary, our \method achieves higher accuracy while preserving communication-efficient characteristics. 

\subsection{Ablation Study}
\begin{table}[ht]
  \caption{The test accuracy (\%) in the practical setting using the HtFE$_8$ model group for ablation study. } 
  \centering
  \resizebox{\linewidth}{!}{
    \begin{tabular}{l|ccc|cc}
    \toprule
     & SCL & FM & w/o $F$ & FedProto & \method \\
     \midrule
     Cifar100 & 40.11 & 43.46 & 40.37 & 36.34 & \textbf{46.94} \\
     Flowers102 & 46.81 & 52.03 & 49.39 & 41.21 & \textbf{53.68} \\
     Tiny-ImageNet & 22.26 & 26.13 & 23.12 & 19.01 & \textbf{27.37} \\
    \bottomrule
    \end{tabular}}
    \label{tab:abl}
\end{table}

We replace \acl with the standard contrastive loss (\cref{eq:standard}), denoted by ``SCL''.
Besides, we modify \acl and \tp by using a fixed margin (\cref{eq:margin})
and removing the further processing model $F$ but only train $\{\acute{P}^c\}^C_{c=1}$, denoted by ``FM'' and ``w/o $F$'', respectively. 
Without utilizing a margin to improve separability, SCL shows a mere improvement of 5.60\% for FedProto on Cifar100, whereas the improvement reaches 10.82\% for FM with a margin. Nevertheless, our adaptive margin can further enhance FM and improve 12.47\% for FedProto on Cifar100. 
Without sufficient trainable parameters in \tp, the performance of w/o $F$ decreases up to 6.57\% compared to our \method, but it still outperforms FedProto by a large gap. 

\subsection{Hyperparameter Study}
\begin{table}[ht]
  \caption{The test accuracy (\%) on Cifar100 in the practical setting using the HtFE$_8$ model group with different $\tau$ or $S$. Recall that we set $\tau=100$ and $S=100$ by default. } 
  \centering
  \resizebox{\linewidth}{!}{
    \begin{tabular}{l|cccc|cccc}
    \toprule
     & \multicolumn{4}{c|}{Different $\tau$} & \multicolumn{4}{c}{Different $S$} \\
     \midrule
     & $1$ & $10$ & $100$ & $1000$ & $1$ & $10$ & $100$ & $1000$ \\
     \midrule
     Acc. & 43.23 & 44.81 & \textbf{46.94} & 46.09 & 43.41 & 44.62 & 46.94 & \textbf{47.01} \\
    \bottomrule
    \end{tabular}}
    \label{tab:hyper}
\end{table}

We evaluate the accuracy of \method by varying the hyperparameters $\tau$ and $S$ in our \method, and the results are shown in \cref{tab:hyper}. Our \method performs better with a larger threshold $\tau$ ranging from $1$ to $100$. However, the accuracy slightly drops when using $\tau=1000$, because an excessively large $\tau$ leads to unstable prototype guidance on clients, and $\delta(t)$ may keep growing during the later stage of training. Unlike $\tau$, increasing the number of server training epochs $S$ leads to higher accuracy in our \method. As the improvement from $S=100$ to $S=1000$ is negligible, we adopt $S=100$ to save computation. Even with $\tau=1$ or $S=1$, our \method can achieve at least 43.23\% in accuracy, which is still higher than baseline methods' accuracy as shown in \cref{tab:datasets_acc} (Practical setting, Cifar100) but setting $S=1$ can save a lot of computation.

\section{Conclusion}

In this work, we propose a novel HtFL method called \method, which shares class-wise prototypes among the server and clients and enhances the separability of different classes via our \tp and \acl. 
Extensive experiments with two statistically heterogeneous settings and twelve heterogeneous models show the superiority of our \method over other baseline methods. 

\section{Acknowledgments}
This work was supported by the National Key R\&D Program of China under Grant No.2022ZD0160504, the Program of Technology Innovation of the Science and Technology Commission of Shanghai Municipality (Granted No. 21511104700), China National Science Foundation (Granted Number 62072301), Tsinghua Toyota Joint Research Institute inter-disciplinary Program, and Tsinghua University(AIR)-Asiainfo Technologies (China) Inc. Joint Research Center.

\bibliography{main}

\begin{thebibliography}{77}
\providecommand{\natexlab}[1]{#1}

\bibitem[{Bakhtiarnia, Zhang, and Iosifidis(2022)}]{bakhtiarnia2022single}
Bakhtiarnia, A.; Zhang, Q.; and Iosifidis, A. 2022.
\newblock Single-layer vision transformers for more accurate early exits with less overhead.
\newblock \emph{Neural Networks}, 153: 461--473.

\bibitem[{Chang et~al.(2023)Chang, Lu, Xue, Xu, and Wei}]{chang2023iterative}
Chang, J.; Lu, Y.; Xue, P.; Xu, Y.; and Wei, Z. 2023.
\newblock Iterative clustering pruning for convolutional neural networks.
\newblock \emph{Knowledge-Based Systems}, 265: 110386.

\bibitem[{Chen and Chao(2021)}]{chen2021bridging}
Chen, H.-Y.; and Chao, W.-L. 2021.
\newblock {On Bridging Generic and Personalized Federated Learning for Image Classification}.
\newblock In \emph{ICLR}.

\bibitem[{Chiang et~al.(2023)Chiang, Frumkin, Liang, and Marculescu}]{chiang2023mobiletl}
Chiang, H.-Y.; Frumkin, N.; Liang, F.; and Marculescu, D. 2023.
\newblock MobileTL: On-Device Transfer Learning with Inverted Residual Blocks.
\newblock In \emph{Proceedings of the AAAI Conference on Artificial Intelligence}.

\bibitem[{Choi, Som, and Turaga(2020)}]{choi2020amc}
Choi, H.; Som, A.; and Turaga, P. 2020.
\newblock AMC-loss: Angular margin contrastive loss for improved explainability in image classification.
\newblock In \emph{CVPR Workshop}.

\bibitem[{Chrabaszcz, Loshchilov, and Hutter(2017)}]{chrabaszcz2017downsampled}
Chrabaszcz, P.; Loshchilov, I.; and Hutter, F. 2017.
\newblock {A Downsampled Variant of Imagenet as an Alternative to the Cifar Datasets}.
\newblock \emph{arXiv preprint arXiv:1707.08819}.

\bibitem[{Collins et~al.(2021)Collins, Hassani, Mokhtari, and Shakkottai}]{collins2021exploiting}
Collins, L.; Hassani, H.; Mokhtari, A.; and Shakkottai, S. 2021.
\newblock {Exploiting Shared Representations for Personalized Federated Learning}.
\newblock In \emph{ICML}.

\bibitem[{Deng et~al.(2019)Deng, Guo, Xue, and Zafeiriou}]{deng2019arcface}
Deng, J.; Guo, J.; Xue, N.; and Zafeiriou, S. 2019.
\newblock Arcface: Additive angular margin loss for deep face recognition.
\newblock In \emph{CVPR}.

\bibitem[{Diao, Ding, and Tarokh(2020)}]{diao2020heterofl}
Diao, E.; Ding, J.; and Tarokh, V. 2020.
\newblock HeteroFL: Computation and Communication Efficient Federated Learning for Heterogeneous Clients.
\newblock In \emph{ICLR}.

\bibitem[{Hayat et~al.(2019)Hayat, Khan, Zamir, Shen, and Shao}]{hayat2019gaussian}
Hayat, M.; Khan, S.; Zamir, S.~W.; Shen, J.; and Shao, L. 2019.
\newblock Gaussian affinity for max-margin class imbalanced learning.
\newblock In \emph{ICCV}.

\bibitem[{He et~al.(2016)He, Zhang, Ren, and Sun}]{he2016deep}
He, K.; Zhang, X.; Ren, S.; and Sun, J. 2016.
\newblock {Deep Residual Learning for Image Recognition}.
\newblock In \emph{CVPR}.

\bibitem[{Hinton, Vinyals, and Dean(2015)}]{hinton2015distilling}
Hinton, G.; Vinyals, O.; and Dean, J. 2015.
\newblock Distilling the knowledge in a neural network.
\newblock \emph{arXiv preprint arXiv:1503.02531}.

\bibitem[{Horvath et~al.(2021)Horvath, Laskaridis, Almeida, Leontiadis, Venieris, and Lane}]{horvath2021fjord}
Horvath, S.; Laskaridis, S.; Almeida, M.; Leontiadis, I.; Venieris, S.; and Lane, N. 2021.
\newblock Fjord: Fair and accurate federated learning under heterogeneous targets with ordered dropout.
\newblock \emph{NeurIPS}.

\bibitem[{Jeong et~al.(2018)Jeong, Oh, Kim, Park, Bennis, and Kim}]{jeong2018communication}
Jeong, E.; Oh, S.; Kim, H.; Park, J.; Bennis, M.; and Kim, S.-L. 2018.
\newblock Communication-efficient on-device machine learning: Federated distillation and augmentation under non-iid private data.
\newblock \emph{arXiv preprint arXiv:1811.11479}.

\bibitem[{Jin, Liu, and Hou(2010)}]{jin2010regularized}
Jin, X.-B.; Liu, C.-L.; and Hou, X. 2010.
\newblock Regularized margin-based conditional log-likelihood loss for prototype learning.
\newblock \emph{Pattern Recognition}, 43(7): 2428--2438.

\bibitem[{Kairouz et~al.(2019)Kairouz, McMahan, Avent, Bellet, Bennis, Bhagoji, Bonawitz, Charles, Cormode, Cummings et~al.}]{kairouz2019advances}
Kairouz, P.; McMahan, H.~B.; Avent, B.; Bellet, A.; Bennis, M.; Bhagoji, A.~N.; Bonawitz, K.; Charles, Z.; Cormode, G.; Cummings, R.; et~al. 2019.
\newblock {Advances and Open Problems in Federated Learning}.
\newblock \emph{arXiv preprint arXiv:1912.04977}.

\bibitem[{Kim and Kim(2020)}]{kim2020attract}
Kim, T.; and Kim, C. 2020.
\newblock Attract, perturb, and explore: Learning a feature alignment network for semi-supervised domain adaptation.
\newblock In \emph{ECCV}.

\bibitem[{Krizhevsky and Geoffrey(2009)}]{krizhevsky2009learning}
Krizhevsky, A.; and Geoffrey, H. 2009.
\newblock {Learning Multiple Layers of Features From Tiny Images}.
\newblock \emph{Technical Report}.

\bibitem[{Krizhevsky, Sutskever, and Hinton(2012)}]{krizhevsky2012imagenet}
Krizhevsky, A.; Sutskever, I.; and Hinton, G.~E. 2012.
\newblock Imagenet classification with deep convolutional neural networks.
\newblock \emph{Advances in neural information processing systems}.

\bibitem[{Leroux et~al.(2018)Leroux, Molchanov, Simoens, Dhoedt, Breuel, and Kautz}]{leroux2018iamnn}
Leroux, S.; Molchanov, P.; Simoens, P.; Dhoedt, B.; Breuel, T.; and Kautz, J. 2018.
\newblock Iamnn: Iterative and adaptive mobile neural network for efficient image classification.
\newblock \emph{arXiv preprint arXiv:1804.10123}.

\bibitem[{Li and Wang(2019)}]{li2019fedmd}
Li, D.; and Wang, J. 2019.
\newblock Fedmd: Heterogenous federated learning via model distillation.
\newblock \emph{arXiv preprint arXiv:1910.03581}.

\bibitem[{Li et~al.(2022{\natexlab{a}})Li, Yue, Wang, Chai, Wang, Tomiyama, and Meng}]{li2022optimizing}
Li, H.; Yue, X.; Wang, Z.; Chai, Z.; Wang, W.; Tomiyama, H.; and Meng, L. 2022{\natexlab{a}}.
\newblock Optimizing the deep neural networks by layer-wise refined pruning and the acceleration on FPGA.
\newblock \emph{Computational Intelligence and Neuroscience}, 2022.

\bibitem[{Li et~al.(2022{\natexlab{b}})Li, Diao, Chen, and He}]{li2022federated}
Li, Q.; Diao, Y.; Chen, Q.; and He, B. 2022{\natexlab{b}}.
\newblock {Federated Learning on Non-IID Data Silos: An Experimental Study}.
\newblock In \emph{ICDE}.

\bibitem[{Li, He, and Song(2021)}]{li2021model}
Li, Q.; He, B.; and Song, D. 2021.
\newblock {Model-Contrastive Federated Learning}.
\newblock In \emph{CVPR}.

\bibitem[{Li et~al.(2021{\natexlab{a}})Li, Wen, Wu, Hu, Wang, Li, Liu, and He}]{li2021survey}
Li, Q.; Wen, Z.; Wu, Z.; Hu, S.; Wang, N.; Li, Y.; Liu, X.; and He, B. 2021{\natexlab{a}}.
\newblock {A Survey on Federated Learning Systems: Vision, Hype and Reality for Data Privacy and Protection}.
\newblock \emph{IEEE Transactions on Knowledge and Data Engineering}.

\bibitem[{Li et~al.(2021{\natexlab{b}})Li, Hu, Beirami, and Smith}]{li2021ditto}
Li, T.; Hu, S.; Beirami, A.; and Smith, V. 2021{\natexlab{b}}.
\newblock {Ditto: Fair and Robust Federated Learning Through Personalization}.
\newblock In \emph{ICML}.

\bibitem[{Li et~al.(2020)Li, Sahu, Talwalkar, and Smith}]{li2020federated}
Li, T.; Sahu, A.~K.; Talwalkar, A.; and Smith, V. 2020.
\newblock {Federated Learning: Challenges, Methods, and Future Directions}.
\newblock \emph{IEEE Signal Processing Magazine}, 37(3): 50--60.

\bibitem[{Li et~al.(2023{\natexlab{a}})Li, Shang, He, Lin, and Wu}]{li2023no}
Li, Z.; Shang, X.; He, R.; Lin, T.; and Wu, C. 2023{\natexlab{a}}.
\newblock No Fear of Classifier Biases: Neural Collapse Inspired Federated Learning with Synthetic and Fixed Classifier.
\newblock \emph{arXiv preprint arXiv:2303.10058}.

\bibitem[{Li et~al.(2023{\natexlab{b}})Li, Wang, Robertson, Clifton, Meinel, and Yang}]{li2023smkd}
Li, Z.; Wang, X.; Robertson, N.~M.; Clifton, D.~A.; Meinel, C.; and Yang, H. 2023{\natexlab{b}}.
\newblock SMKD: Selective Mutual Knowledge Distillation.
\newblock In \emph{IJCNN}.

\bibitem[{Liang et~al.(2020)Liang, Liu, Ziyin, Allen, Auerbach, Brent, Salakhutdinov, and Morency}]{liang2020think}
Liang, P.~P.; Liu, T.; Ziyin, L.; Allen, N.~B.; Auerbach, R.~P.; Brent, D.; Salakhutdinov, R.; and Morency, L.-P. 2020.
\newblock Think locally, act globally: Federated learning with local and global representations.
\newblock \emph{arXiv preprint arXiv:2001.01523}.

\bibitem[{Liao et~al.(2023)Liao, Ma, Zhou, Zhao, and Xie}]{liao2023draftfed}
Liao, Y.; Ma, L.; Zhou, B.; Zhao, X.; and Xie, F. 2023.
\newblock DraftFed: A Draft-Based Personalized Federated Learning Approach for Heterogeneous Convolutional Neural Networks.
\newblock \emph{IEEE Transactions on Mobile Computing}.

\bibitem[{Lin et~al.(2022)Lin, Ji, Ji, and Yao}]{lin2022closer}
Lin, S.; Ji, B.; Ji, R.; and Yao, A. 2022.
\newblock A closer look at branch classifiers of multi-exit architectures.
\newblock \emph{arXiv preprint arXiv:2204.13347}.

\bibitem[{Lin et~al.(2020)Lin, Kong, Stich, and Jaggi}]{lin2020ensemble}
Lin, T.; Kong, L.; Stich, S.~U.; and Jaggi, M. 2020.
\newblock Ensemble distillation for robust model fusion in federated learning.
\newblock \emph{NeurIPS}.

\bibitem[{Luo et~al.(2021)Luo, Chen, Hu, Zhang, Liang, and Feng}]{luo2021no}
Luo, M.; Chen, F.; Hu, D.; Zhang, Y.; Liang, J.; and Feng, J. 2021.
\newblock {No Fear of Heterogeneity: Classifier Calibration for Federated Learning with Non-IID data}.
\newblock In \emph{NeurIPS}.

\bibitem[{Ma et~al.(2022)Ma, Zhang, Guo, and Xu}]{ma2022layer}
Ma, X.; Zhang, J.; Guo, S.; and Xu, W. 2022.
\newblock Layer-wised model aggregation for personalized federated learning.
\newblock In \emph{CVPR}.

\bibitem[{McMahan et~al.(2017)McMahan, Moore, Ramage, Hampson, and y~Arcas}]{mcmahan2017communication}
McMahan, B.; Moore, E.; Ramage, D.; Hampson, S.; and y~Arcas, B.~A. 2017.
\newblock {Communication-Efficient Learning of Deep Networks from Decentralized Data}.
\newblock In \emph{AISTATS}.

\bibitem[{Nakatsukasa and Higham(2013)}]{nakatsukasa2013stable}
Nakatsukasa, Y.; and Higham, N.~J. 2013.
\newblock Stable and efficient spectral divide and conquer algorithms for the symmetric eigenvalue decomposition and the SVD.
\newblock \emph{SIAM Journal on Scientific Computing}, 35(3): A1325--A1349.

\bibitem[{Nilsback and Zisserman(2008)}]{nilsback2008automated}
Nilsback, M.-E.; and Zisserman, A. 2008.
\newblock Automated flower classification over a large number of classes.
\newblock In \emph{2008 Sixth Indian conference on computer vision, graphics \& image processing}, 722--729. IEEE.

\bibitem[{Pinheiro(2018)}]{pinheiro2018unsupervised}
Pinheiro, P.~O. 2018.
\newblock Unsupervised domain adaptation with similarity learning.
\newblock In \emph{CVPR}.

\bibitem[{Sandler et~al.(2018)Sandler, Howard, Zhu, Zhmoginov, and Chen}]{sandler2018mobilenetv2}
Sandler, M.; Howard, A.; Zhu, M.; Zhmoginov, A.; and Chen, L.-C. 2018.
\newblock Mobilenetv2: Inverted residuals and linear bottlenecks.
\newblock In \emph{CVPR}.

\bibitem[{Schroff, Kalenichenko, and Philbin(2015)}]{schroff2015facenet}
Schroff, F.; Kalenichenko, D.; and Philbin, J. 2015.
\newblock Facenet: A unified embedding for face recognition and clustering.
\newblock In \emph{CVPR}.

\bibitem[{Shamsian et~al.(2021)Shamsian, Navon, Fetaya, and Chechik}]{shamsian2021personalized}
Shamsian, A.; Navon, A.; Fetaya, E.; and Chechik, G. 2021.
\newblock Personalized federated learning using hypernetworks.
\newblock In \emph{ICML}.

\bibitem[{Shen et~al.(2020)Shen, Zhang, Jia, Zhang, Huang, Zhou, Kuang, Wu, and Wu}]{shen2020federated}
Shen, T.; Zhang, J.; Jia, X.; Zhang, F.; Huang, G.; Zhou, P.; Kuang, K.; Wu, F.; and Wu, C. 2020.
\newblock Federated mutual learning.
\newblock \emph{arXiv preprint arXiv:2006.16765}.

\bibitem[{Shin et~al.(2023)Shin, Kwak, Kim, Ramstr{\"o}m, Jeong, Ha, and Kim}]{shin2023scaling}
Shin, K.; Kwak, H.; Kim, S.~Y.; Ramstr{\"o}m, M.~N.; Jeong, J.; Ha, J.-W.; and Kim, K.-M. 2023.
\newblock Scaling law for recommendation models: Towards general-purpose user representations.
\newblock In \emph{AAAI}.

\bibitem[{Szegedy et~al.(2015)Szegedy, Liu, Jia, Sermanet, Reed, Anguelov, Erhan, Vanhoucke, and Rabinovich}]{szegedy2015going}
Szegedy, C.; Liu, W.; Jia, Y.; Sermanet, P.; Reed, S.; Anguelov, D.; Erhan, D.; Vanhoucke, V.; and Rabinovich, A. 2015.
\newblock Going deeper with convolutions.
\newblock In \emph{CVPR}.

\bibitem[{T~Dinh, Tran, and Nguyen(2020)}]{t2020personalized}
T~Dinh, C.; Tran, N.; and Nguyen, T.~D. 2020.
\newblock {Personalized Federated Learning with Moreau Envelopes}.
\newblock In \emph{NeurIPS}.

\bibitem[{Tan et~al.(2022{\natexlab{a}})Tan, Yu, Cui, and Yang}]{tan2022towards}
Tan, A.~Z.; Yu, H.; Cui, L.; and Yang, Q. 2022{\natexlab{a}}.
\newblock {Towards Personalized Federated Learning}.
\newblock \emph{IEEE Transactions on Neural Networks and Learning Systems}.
\newblock Early Access.

\bibitem[{Tan et~al.(2022{\natexlab{b}})Tan, Long, Liu, Zhou, Lu, Jiang, and Zhang}]{tan2022fedproto}
Tan, Y.; Long, G.; Liu, L.; Zhou, T.; Lu, Q.; Jiang, J.; and Zhang, C. 2022{\natexlab{b}}.
\newblock {Fedproto: Federated Prototype Learning across Heterogeneous Clients}.
\newblock In \emph{AAAI}.

\bibitem[{Tan et~al.(2022{\natexlab{c}})Tan, Long, Ma, Liu, Zhou, and Jiang}]{tan2022federated}
Tan, Y.; Long, G.; Ma, J.; Liu, L.; Zhou, T.; and Jiang, J. 2022{\natexlab{c}}.
\newblock Federated Learning from Pre-Trained Models: A Contrastive Learning Approach.
\newblock \emph{arXiv preprint arXiv:2209.10083}.

\bibitem[{Tanwisuth et~al.(2021)Tanwisuth, Fan, Zheng, Zhang, Zhang, Chen, and Zhou}]{tanwisuth2021prototype}
Tanwisuth, K.; Fan, X.; Zheng, H.; Zhang, S.; Zhang, H.; Chen, B.; and Zhou, M. 2021.
\newblock A prototype-oriented framework for unsupervised domain adaptation.
\newblock \emph{NeurIPS}.

\bibitem[{Van~der Maaten and Hinton(2008)}]{van2008visualizing}
Van~der Maaten, L.; and Hinton, G. 2008.
\newblock {Visualizing Data Using T-SNE}.
\newblock \emph{Journal of Machine Learning Research}, 9(11).

\bibitem[{Wang et~al.(2020)Wang, Yurochkin, Sun, Papailiopoulos, and Khazaeni}]{wang2020federated}
Wang, H.; Yurochkin, M.; Sun, Y.; Papailiopoulos, D.; and Khazaeni, Y. 2020.
\newblock Federated learning with matched averaging.
\newblock \emph{arXiv preprint arXiv:2002.06440}.

\bibitem[{Wang et~al.(2023)Wang, Wang, Zhang, and Fu}]{wang2023model}
Wang, L.; Wang, M.; Zhang, D.; and Fu, H. 2023.
\newblock Model Barrier: A Compact Un-Transferable Isolation Domain for Model Intellectual Property Protection.
\newblock In \emph{CVPR}.

\bibitem[{Wen, Jeon, and Huang(2022)}]{wen2022federated}
Wen, D.; Jeon, K.-J.; and Huang, K. 2022.
\newblock Federated dropout—A simple approach for enabling federated learning on resource constrained devices.
\newblock \emph{IEEE wireless communications letters}, 11(5): 923--927.

\bibitem[{Wu et~al.(2022)Wu, Wu, Lyu, Huang, and Xie}]{wu2022communication}
Wu, C.; Wu, F.; Lyu, L.; Huang, Y.; and Xie, X. 2022.
\newblock Communication-efficient federated learning via knowledge distillation.
\newblock \emph{Nature communications}, 13(1): 2032.

\bibitem[{Xiao, Rasul, and Vollgraf(2017)}]{xiao2017fashion}
Xiao, H.; Rasul, K.; and Vollgraf, R. 2017.
\newblock {Fashion-MNIST: A Novel Image Dataset for Benchmarking Machine Learning Algorithms}.
\newblock \emph{arXiv preprint arXiv:1708.07747}.

\bibitem[{Xu et~al.(2020)Xu, Xian, Wang, Schiele, and Akata}]{xu2020attribute}
Xu, W.; Xian, Y.; Wang, J.; Schiele, B.; and Akata, Z. 2020.
\newblock Attribute prototype network for zero-shot learning.
\newblock \emph{NeurIPS}.

\bibitem[{Yan, Wang, and Li(2022)}]{yan2022seizing}
Yan, G.; Wang, H.; and Li, J. 2022.
\newblock Seizing critical learning periods in federated learning.
\newblock In \emph{Proceedings of the AAAI Conference on Artificial Intelligence}.

\bibitem[{Yang et~al.(2018)Yang, Zhang, Yin, and Liu}]{yang2018robust}
Yang, H.-M.; Zhang, X.-Y.; Yin, F.; and Liu, C.-L. 2018.
\newblock Robust classification with convolutional prototype learning.
\newblock In \emph{CVPR}.

\bibitem[{Yang, Huang, and Ye(2023)}]{yang2023dynamic}
Yang, X.; Huang, W.; and Ye, M. 2023.
\newblock Dynamic Personalized Federated Learning with Adaptive Differential Privacy.
\newblock In \emph{NeurIPS}.

\bibitem[{Yi et~al.(2023)Yi, Wang, Liu, Shi, and Yu}]{yi2023fedgh}
Yi, L.; Wang, G.; Liu, X.; Shi, Z.; and Yu, H. 2023.
\newblock FedGH: Heterogeneous Federated Learning with Generalized Global Header.
\newblock \emph{arXiv preprint arXiv:2303.13137}.

\bibitem[{Yu et~al.(2022)Yu, Liu, Wang, Xu, and Liu}]{yu2022multimodal}
Yu, Q.; Liu, Y.; Wang, Y.; Xu, K.; and Liu, J. 2022.
\newblock Multimodal Federated Learning via Contrastive Representation Ensemble.
\newblock In \emph{ICLR}.

\bibitem[{Zhang et~al.(2018{\natexlab{a}})Zhang, Gu, Jang, Wu, Stoecklin, Huang, and Molloy}]{zhang2018protecting}
Zhang, J.; Gu, Z.; Jang, J.; Wu, H.; Stoecklin, M.~P.; Huang, H.; and Molloy, I. 2018{\natexlab{a}}.
\newblock Protecting intellectual property of deep neural networks with watermarking.
\newblock In \emph{ASIA-CCS}.

\bibitem[{Zhang et~al.(2023{\natexlab{a}})Zhang, Guo, Guo, Zeng, Zhou, and Zomaya}]{zhang2023towards}
Zhang, J.; Guo, S.; Guo, J.; Zeng, D.; Zhou, J.; and Zomaya, A. 2023{\natexlab{a}}.
\newblock Towards Data-Independent Knowledge Transfer in Model-Heterogeneous Federated Learning.
\newblock \emph{IEEE Transactions on Computers}.

\bibitem[{Zhang et~al.(2021)Zhang, Guo, Ma, Wang, Xu, and Wu}]{zhang2021parameterized}
Zhang, J.; Guo, S.; Ma, X.; Wang, H.; Xu, W.; and Wu, F. 2021.
\newblock {Parameterized Knowledge Transfer for Personalized Federated Learning}.
\newblock In \emph{NeurIPS}.

\bibitem[{Zhang et~al.(2023{\natexlab{b}})Zhang, Hua, Cao, Wang, Song, XUE, Ma, and Guan}]{zhang2023eliminating}
Zhang, J.; Hua, Y.; Cao, J.; Wang, H.; Song, T.; XUE, Z.; Ma, R.; and Guan, H. 2023{\natexlab{b}}.
\newblock Eliminating Domain Bias for Federated Learning in Representation Space.
\newblock In \emph{NeurIPS}.

\bibitem[{Zhang et~al.(2023{\natexlab{c}})Zhang, Hua, Wang, Song, Xue, Ma, Cao, and Guan}]{zhang2023gpfl}
Zhang, J.; Hua, Y.; Wang, H.; Song, T.; Xue, Z.; Ma, R.; Cao, J.; and Guan, H. 2023{\natexlab{c}}.
\newblock GPFL: Simultaneously Learning Global and Personalized Feature Information for Personalized Federated Learning.
\newblock In \emph{ICCV}.

\bibitem[{Zhang et~al.(2023{\natexlab{d}})Zhang, Hua, Wang, Song, Xue, Ma, and Guan}]{zhang2022fedala}
Zhang, J.; Hua, Y.; Wang, H.; Song, T.; Xue, Z.; Ma, R.; and Guan, H. 2023{\natexlab{d}}.
\newblock {FedALA: Adaptive Local Aggregation for Personalized Federated Learning}.
\newblock In \emph{AAAI}.

\bibitem[{Zhang et~al.(2023{\natexlab{e}})Zhang, Hua, Wang, Song, Xue, Ma, and Guan}]{Zhang2023fedcp}
Zhang, J.; Hua, Y.; Wang, H.; Song, T.; Xue, Z.; Ma, R.; and Guan, H. 2023{\natexlab{e}}.
\newblock FedCP: Separating Feature Information for Personalized Federated Learning via Conditional Policy.
\newblock In \emph{KDD}.

\bibitem[{Zhang and Sato(2023)}]{zhang2023semantic}
Zhang, K.; and Sato, Y. 2023.
\newblock Semantic Image Segmentation by Dynamic Discriminative Prototypes.
\newblock \emph{IEEE Transactions on Multimedia}.

\bibitem[{Zhang et~al.(2022)Zhang, Shen, Ding, Tao, and Duan}]{zhang2022fine}
Zhang, L.; Shen, L.; Ding, L.; Tao, D.; and Duan, L.-Y. 2022.
\newblock {Fine-Tuning Global Model Via Data-Free Knowledge Distillation for Non-IID Federated Learning}.
\newblock In \emph{CVPR}.

\bibitem[{Zhang et~al.(2018{\natexlab{b}})Zhang, Xiang, Hospedales, and Lu}]{zhang2018deep}
Zhang, Y.; Xiang, T.; Hospedales, T.~M.; and Lu, H. 2018{\natexlab{b}}.
\newblock Deep mutual learning.
\newblock In \emph{CVPR}.

\bibitem[{Zhao and Wang(2022)}]{zhao2022new}
Zhao, L.; and Wang, L. 2022.
\newblock A new lightweight network based on MobileNetV3.
\newblock \emph{KSII Transactions on Internet \& Information Systems}, 16(1).

\bibitem[{Zhao et~al.(2022)Zhao, Wang, Jia, and Cui}]{zhao2022lightweight}
Zhao, L.; Wang, L.; Jia, Y.; and Cui, Y. 2022.
\newblock A lightweight deep neural network with higher accuracy.
\newblock \emph{Plos one}, 17(8): e0271225.

\bibitem[{Zhong et~al.(2017)Zhong, Li, Ma, Jiang, and Zhao}]{zhong2017deep}
Zhong, Z.; Li, J.; Ma, L.; Jiang, H.; and Zhao, H. 2017.
\newblock Deep residual networks for hyperspectral image classification.
\newblock In \emph{IEEE international geoscience and remote sensing symposium (IGARSS)}.

\bibitem[{Zhu, Hong, and Zhou(2021)}]{zhu2021data}
Zhu, Z.; Hong, J.; and Zhou, J. 2021.
\newblock {Data-Free Knowledge Distillation for Heterogeneous Federated Learning}.
\newblock In \emph{ICML}.

\bibitem[{Zhuang, Chen, and Lyu(2023)}]{zhuang2023foundation}
Zhuang, W.; Chen, C.; and Lyu, L. 2023.
\newblock When Foundation Model Meets Federated Learning: Motivations, Challenges, and Future Directions.
\newblock \emph{arXiv preprint arXiv:2306.15546}.

\end{thebibliography}

\appendix

\section{Additional Experimental Details}

\noindent\textbf{Experimental environment. \ } 
All our experiments are conducted on a machine with 64 Intel(R) Xeon(R) Platinum 8362 CPUs, 256G memory, eight NVIDIA 3090 GPUs, and Ubuntu 20.04.4 LTS. 

\noindent\textbf{Hyperparameter settings. \ } 
In addition to the hyperparameter settings provided in the main body, we adhere to each baseline method's original paper for their respective hyperparameter settings.
LG-FedAvg~\cite{liang2020think} has no additional hyperparameters. 
For FedGen~\cite{zhu2021data}, we set the noise dimension to 32, its generator learning rate to 0.1, its hidden dimension to be equal to the feature dimension, \ie, $K$, and its server learning epochs to 100. For FML~\cite{shen2020federated}, we set its knowledge distillation hyperparameters $\alpha=0.5$, and $\beta=0.5$. For FedKD~\cite{wu2022communication}, we set its auxiliary model learning rate to be the same as the one of the client model, \ie, 0.01, $T_{start}=0.95$ and $T_{end}=0.95$. For FedDistill~\cite{jeong2018communication}, we set $\gamma=1$. For FedProto~\cite{tan2022fedproto}, we set $\lambda=0.1$. For our \method, we set $\lambda=0.1$, margin threshold $\tau=100$, and server learning epochs $S=100$. We use the same hyperparameter settings for all the tasks. 

\noindent\textbf{Model heterogeneity. \ } 
Since FedGen and LG-FedAvg require homogeneous classifiers for parameter aggregation on the server by default, we consider the last FC layer (homogeneous) and the rest of the layers (heterogeneous) as the classifier and feature extractors for all methods, respectively, by default. Furthermore, we also consider heterogeneous classifiers (\ie, HtC$_4$) and heterogeneous extractors for more general scenarios involving various model structures. 
According to FedKD and FML, the auxiliary model needs to be designed as small as possible to reduce the communication overhead for model parameter transmitting, so we choose the smallest model in any given model group to be the auxiliary model for FedKD and FML. Specifically, we choose the 4-layer CNN as the auxiliary model for the scenarios that use homogeneous models in \cref{tab:homo}. 

\noindent\textbf{Architectures of HtC$_4$. \ } 
We construct the HtC$_4$ model group for Cifar100 (100 classes) in Tab. 2 of the main body using four different classifier architectures consisting solely of FC layers. Following the notations in \citet{he2016deep}, we present these architectures as follows: 
\begin{enumerate}
    \item 100-d fc: This architecture consists of a single 100-way FC layer.
    \item 512-d fc, 100-d fc: This architecture includes two FC layers connected sequentially. These two FC layers are 512-way and 100-way, respectively. 
    \item 256-d fc, 100-d fc: This architecture includes two FC layers connected sequentially. These two FC layers are 256-way and 100-way, respectively. 
    \item 128-d fc, 100-d fc: This architecture includes two FC layers connected sequentially. These two FC layers are 128-way and 100-way, respectively. 
\end{enumerate}

\noindent\textbf{FLOPs computing. \ } 
To estimate the number of floating-point operations (FLOPs) for each HtFL method, we only consider the operations performed during the forward and backward passes involving the trainable parameters. Other operations, such as data preprocessing, are not included in the FLOPs calculation. According to prior work~\cite{chiang2023mobiletl}, the backward pass requires approximately double the FLOPs of the forward pass. We list the FLOPs of our considered model architectures in \cref{tab:flop}. Then, we can obtain the total FLOPs per iteration across active clients by multiplying the number of active clients with the tripled FLOPs of the forward pass. 
\begin{table*}[ht]
  \caption{The forward FLOPs of the architectures in the HtFE$_8$ model group on Cifar100. ``B'' is short for billion. } 
  \centering
  \resizebox{!}{!}{
    \begin{tabular}{l|r|l}
    \toprule
     & FLOPs & References\\
     \midrule
     4-layer CNN & 0.013B & None \\
     GoogleNet & 1.530B & \citet{chang2023iterative, lin2022closer} \\
     MobileNet\_v2 & 0.314B & \citet{zhao2022lightweight} \\
     ResNet18 & 0.117B & \citet{zhao2022new} \\
     ResNet34 & 0.218B & \citet{zhao2022new} \\
     ResNet50 & 1.305B & \citet{li2022optimizing} \\
     ResNet101 & 2.532B & \citet{li2022optimizing, leroux2018iamnn} \\
     ResNet152 & 5.330B & \citet{bakhtiarnia2022single} \\
    \bottomrule
    \end{tabular}}
    \label{tab:flop}
\end{table*}

\section{Additional Experimental Results}

In addition to the extensive experiments presented in the main body, we also conducted additional comparison experiments to further evaluate the effectiveness of our \method. 

\subsection{Performance on Fashion-MNIST}

\begin{table*}[ht]
  \caption{The model architectures in HtCNN$_8$ model group. We follow \citet{he2016deep} to denote the convolutional layer~\cite{krizhevsky2012imagenet} and the pooling layer. For example, ``[$5\times 5, 32$]'' represents a convolutional layer with kernel size $5\times 5$ and output channel $32$ while ``$2\times2$ max pool'' represents a max pooling layer with kernel size $2\times 2$. }
  \centering
  \resizebox{!}{!}{
    \begin{tabular}{l|ll}
    \toprule
     & Sequentially Connected Feature Extractors & Classifiers \\
    \midrule
    CNN1 & [$5\times 5, 32$], $2\times2$ max pool, 512-d fc & 10-d fc \\
    CNN2 & [$5\times 5, 32$], $2\times2$ max pool, [$5\times 5, 64$], $2\times2$ max pool, 512-d fc & 10-d fc \\
    CNN3 & [$5\times 5, 32$], $2\times2$ max pool, 512-d fc, 512-d fc & 10-d fc \\
    CNN4 & [$5\times 5, 32$], $2\times2$ max pool, [$5\times 5, 64$], $2\times2$ max pool, 512-d fc, 512-d fc & 10-d fc \\
    CNN5 & [$5\times 5, 32$], $2\times2$ max pool, 1024-d fc, 512-d fc & 10-d fc \\
    CNN6 & [$5\times 5, 32$], $2\times2$ max pool, [$5\times 5, 64$], $2\times2$ max pool, 1024-d fc, 512-d fc & 10-d fc \\
    CNN7 & [$5\times 5, 32$], $2\times2$ max pool, 1024-d fc, 512-d fc, 512-d fc & 10-d fc \\
    CNN8 & [$5\times 5, 32$], $2\times2$ max pool, [$5\times 5, 64$], $2\times2$ max pool, 1024-d fc, 512-d fc, 512-d fc & 10-d fc \\
    \bottomrule
    \end{tabular}}
    \label{tab:HtCNNs}
\end{table*}

\begin{table}[ht]
  \caption{The test accuracy (\%) on the FMNIST dataset using the HtCNN$_8$ model group. }
  \centering
  \resizebox{\linewidth}{!}{
    \begin{tabular}{l|cc}
    \toprule
    Settings & Pathological Setting & Practical Setting \\
    \midrule
    LG-FedAvg & 99.39$\pm$0.01 & 97.23$\pm$0.03 \\
    FedGen & 99.38$\pm$0.04 & 97.35$\pm$0.02 \\
    FML & 99.42$\pm$0.02 & 97.36$\pm$0.03 \\
    FedKD & 99.37$\pm$0.06 & 97.30$\pm$0.04 \\
    FedDistill & 99.47$\pm$0.01 & 97.48$\pm$0.04 \\
    FedProto & 99.48$\pm$0.01 & 97.46$\pm$0.01 \\
    \midrule
    \method & \textbf{99.56$\pm$0.03} & \textbf{97.58$\pm$0.05} \\
    \bottomrule
    \end{tabular}}
    \label{tab:fmnist}
\end{table}

We also evaluate our \method on another popular dataset Fashion-MNIST (FMNIST)~\cite{xiao2017fashion} in both the pathological and practical settings. Specifically, we assign non-redundant and unbalanced data of 2 classes to each client from a total of 10 classes on FMNIST and use the default practical setting (\ie, $\beta=0.1$). Since each image in FMNIST is a grayscale image that only contains one channel, all the model architectures adopted in the main body are not applicable here. Therefore, we create another model group that contains eight model architectures, called HtCNN$_8$, as listed in \cref{tab:HtCNNs}. We allocate these architectures to clients using the approach introduced in HtFE$_X$. According to \cref{tab:fmnist}, our \method can also outperform other baselines on FMNIST. 

\subsection{Homogeneous Models}
\label{sec:homo}

\begin{table}[ht]
  \caption{The test accuracy (\%) on Cifar100 in the practical setting using homogeneous models (identical architectures). }
  \centering
  \resizebox{\linewidth}{!}{
    \begin{tabular}{l|ccc}
    \toprule
    Architectures & ResNet10 & ResNet18 & ResNet34 \\
    \midrule
    LG-FedAvg & 47.27$\pm$0.22 & 44.74$\pm$0.16 & 44.24$\pm$0.07 \\
    FedGen & 46.42$\pm$0.13 & 44.05$\pm$0.05 & 43.71$\pm$0.04 \\
    FML & 46.71$\pm$0.08 & 42.91$\pm$0.27 & 40.21$\pm$0.18 \\
    FedKD & 45.29$\pm$0.12 & 41.13$\pm$0.13 & 39.85$\pm$0.21 \\
    FedDistill & 44.54$\pm$0.14 & 43.83$\pm$0.22 & 43.31$\pm$0.19 \\
    FedProto & 40.15$\pm$0.60 & 39.91$\pm$0.03 & 37.22$\pm$0.13 \\
    \midrule
    \method & \textbf{49.40$\pm$0.51} & \textbf{46.47$\pm$0.96} & \textbf{46.42$\pm$0.95} \\
    \bottomrule
    \end{tabular}}
    \label{tab:homo}
\end{table}

Here we remove the model heterogeneity by using homogeneous models for all clients. Thus, only statistical heterogeneity exists in these scenarios. The results are shown in \cref{tab:homo}, where our \method still outperforms other methods. Without sharing the feature extractor part, all the methods perform worse with larger models due to local data scarcity. 
The performance of the global classifier (LG-FedAvg and FedGen), auxiliary model (FML and FedKD), and global prototypes (FedDistill, FedProto, and our \method) heavily relies on the private feature extractors. However, when using larger models with deeper feature extractors, the feature extractor training becomes challenging, especially in early iterations. This can lead to suboptimal global classifiers, auxiliary models, or global prototypes, which in turn negatively impact the training of the feature extractors in subsequent iterations. Consequently, this iterative process can result in lower accuracy overall, as the training in early iterations is critical in FL~\cite{yan2022seizing}. However, our \method only drops 0.05\% in accuracy from using ResNet18 to using ResNet34, while baselines drop around 0.34\%$\sim$2.70\%. 

\subsection{Computation Cost}

\begin{table}[ht]
  \caption{The total FLOPs on clients per iteration using the HtFE$_8$ model group on Cifar100 in the practical setting. ``B'' is short for billion. The symbol $^\dag$ denotes that the cost of SVD is not included. }
  \setlength{\tabcolsep}{3pt}
  \centering
  \resizebox{!}{!}{
    \begin{tabular}{l|c}
    \toprule
    & Computation Cost \\
    \midrule
    LG-FedAvg & 98.77B \\
    FedGen & 98.78B \\
    FML & 99.81B \\
    FedKD & 99.81B$^\dag$ \\
    FedDistill & 98.77B \\
    FedProto & 98.77B \\
    \midrule
    \method & 98.77B \\
    \bottomrule
    \end{tabular}}
    \label{tab:comp}
\end{table}

We show the total (approximated) computation cost on all clients per iteration and report the FLOPs in \cref{tab:comp}. Note that the computation cost of clients is often considered a challenging bottleneck in FL, especially for resource-constrained edge devices, while the computation power of the server is often assumed to be abundant in FL~\cite{kairouz2019advances, zhang2022fine}. 
According to \cref{tab:comp}, we show that although all methods cost comparable computation overhead, FML and FedKD require an additional 1.04 billion FLOP introduced by the auxiliary model, which is considerable. 
Although FedKD reduces the communication overhead through singular value decomposition (SVD) on the auxiliary model parameters before uploading to the server, it additionally costs 9.35B FLOPs for clients to compute per iteration~\cite{nakatsukasa2013stable}. 
Although our \method incurs the same computation overhead on the clients per iteration as FedProto, it achieves a significant improvement in efficiency. Specifically, \method only requires 17 iterations (with a total time of 80.47 minutes) to achieve an accuracy of 36.34\%, whereas FedProto takes 489 iterations (with a total time of 1613.70 minutes) to reach the same accuracy on the same machine.

\section{Visualizations}

\subsection{Training Error Curve}

\begin{figure}[h]
	\centering
	\includegraphics[width=\linewidth]{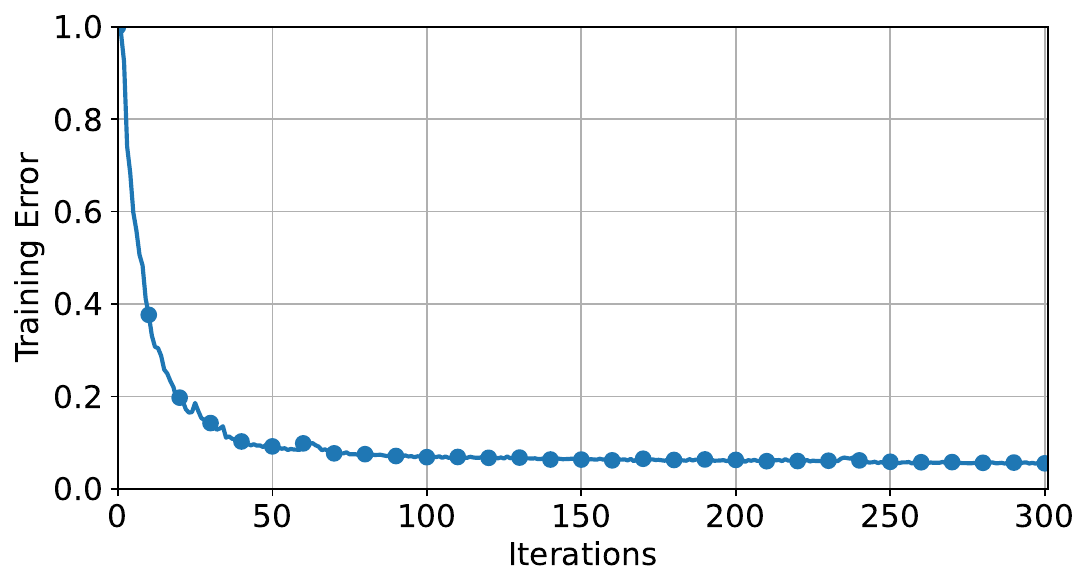}
	\caption{The training error curve on Flowers102 using the HtFE$_8$ model group in the default practical setting.}
    \label{fig:converge}
\end{figure}

We show the training error curve of our \method in \cref{fig:converge}, where we calculate the training error on clients' training sets in the same way as calculating test accuracy in the main body. According to \cref{fig:converge}, our \method optimizes quickly in the initial 50 iterations and gradually converges in the subsequent iterations. Besides, our \method maintains stable performance after converging at around the 100th iteration. 

\subsection{Visualizations of Prototypes}

\begin{figure*}[ht]
	\centering
	\subfigure[FedProto]{\includegraphics[width=0.48\linewidth]{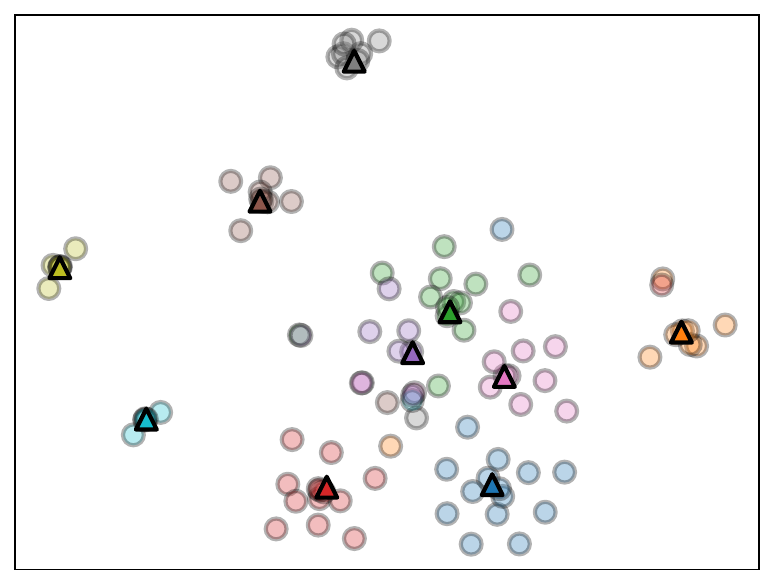}\label{fig:tsne_proto_fedproto}}
    \hfill
	\subfigure[\method]{\includegraphics[width=0.48\linewidth]{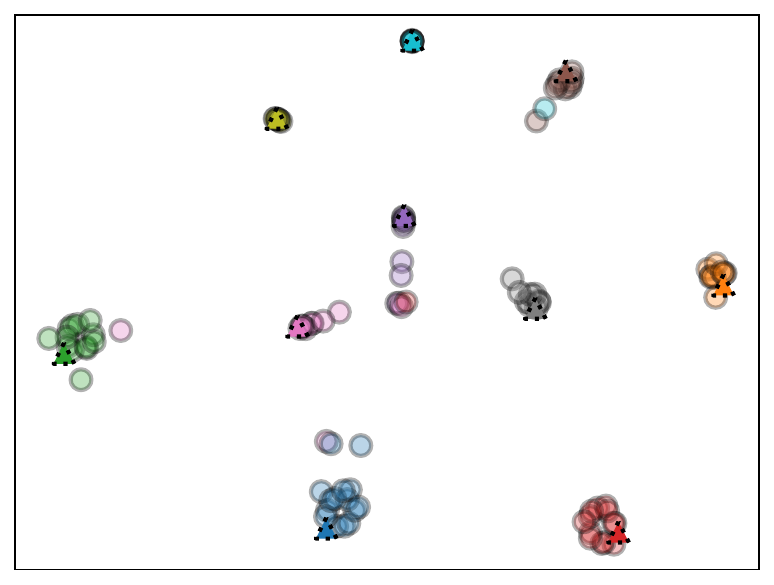}\label{fig:tsne_proto_our}}
	\caption{The t-SNE visualization of prototypes on the server on FMNIST in the practical setting using the HtCNN$_8$ model group. Different colors represent different classes. Circles represent the client prototypes and triangles represent the global prototypes. Triangles with dotted borders represent our \tp. \textit{Best viewed in color.} } \label{fig:tsne_proto}
\end{figure*}

In the main body, we have shown a symbolic figure, \ie, Fig. 2 (main body), to illustrate the mechanism of our key component \tp on the server. Here, we borrow the icons in Fig. 2 (main body) to demonstrate a t-SNE~\cite{van2008visualizing} visualization of prototypes on the experimental data when FedProto and our \method have converged, as shown in \cref{fig:tsne_proto}. According to the \textit{triangles} in \cref{fig:tsne_proto}, we observe that the weighted-averaging in FedProto generates global prototypes with smaller prototype margins than the best-separated client prototypes, while our \method can push global prototypes away from each other to retain the maximum prototype margin. Meanwhile, it is worth noting that \method maintains the semantics of the prototypes. This is evident as the new global prototypes generated by our method remain within the range of client prototypes. Guided by our separable global prototypes, clients' heterogeneous feature extractors can generate more compact and discriminative client prototypes in \method than FedProto, as shown by the \textit{circles} in \cref{fig:tsne_proto}. 

\subsection{Visualizations of Feature Representations}

\begin{figure*}[ht]
	\centering
	\subfigure[FedProto]{\includegraphics[width=0.48\linewidth]{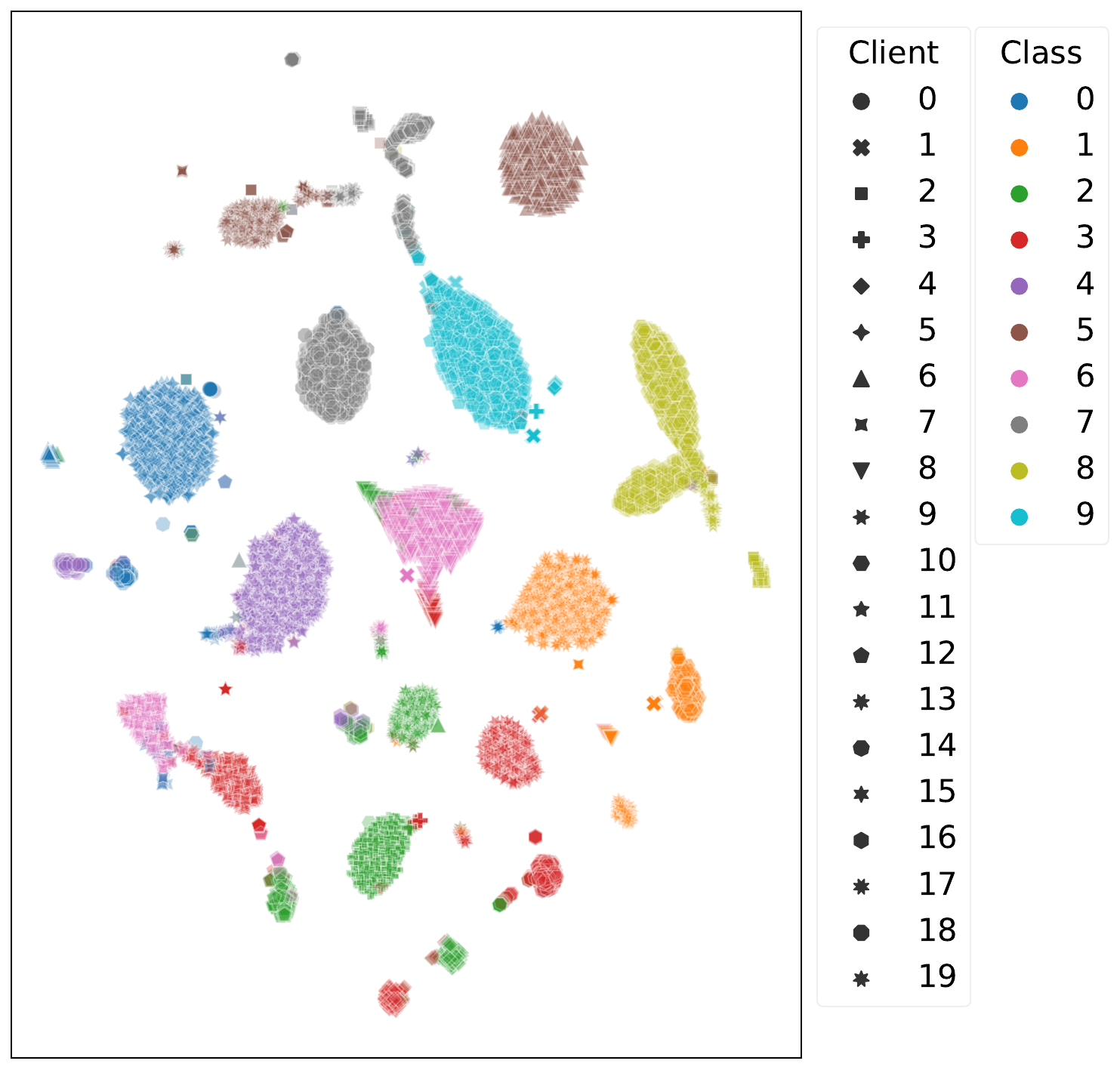}\label{fig:tsne_rep_fedproto}}
    \hfill
	\subfigure[\method]{\includegraphics[width=0.48\linewidth]{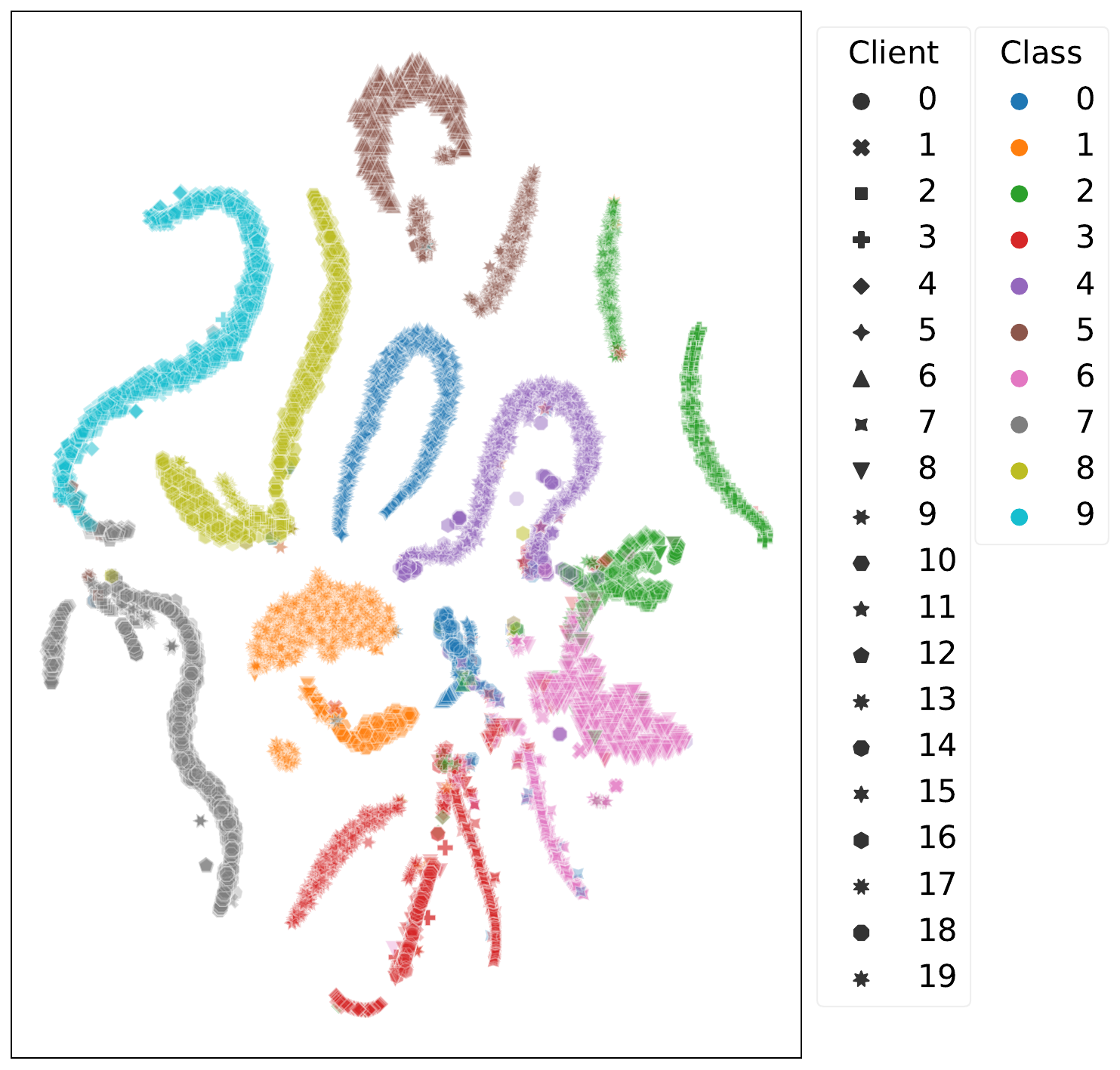}\label{fig:tsne_rep_our}}
	\caption{The t-SNE visualization of the feature representations on all clients' test sets on FMNIST in the practical setting using the HtCNN$_8$ model group. \textit{Best viewed in color.} } \label{fig:tsne_rep}
\end{figure*}

Besides the t-SNE visualization of prototypes on the server, we also illustrate the t-SNE visualization of the feature representations on all clients' test sets in \cref{fig:tsne_rep} when FedProto and our \method have converged. Based on \cref{fig:tsne_rep}, we find that the feature representations of different classes overlap or mix in FedProto since it guides client model training by its informative global prototypes. In contrast, our \method guides clients' model training with separable global prototypes, as shown in \cref{fig:tsne_proto}. Thus, clients' models can extract discriminate feature representations, which further provide high-quality client prototypes to facilitate our \tp learning on the server. Moreover, the presence of model heterogeneity makes it challenging for the feature representations of the same class from different clients to cluster together in the t-SNE visualization. However, our \method can cluster these feature representations closer together compared to FedProto. 

\subsection{Visualizations of Data Distributions}

We illustrate the data distributions (including training and test data) in the experiments here. 

\begin{figure*}[ht]
	\centering
	\hfill
	\subfigure[FMNIST]{\includegraphics[width=0.32\linewidth]{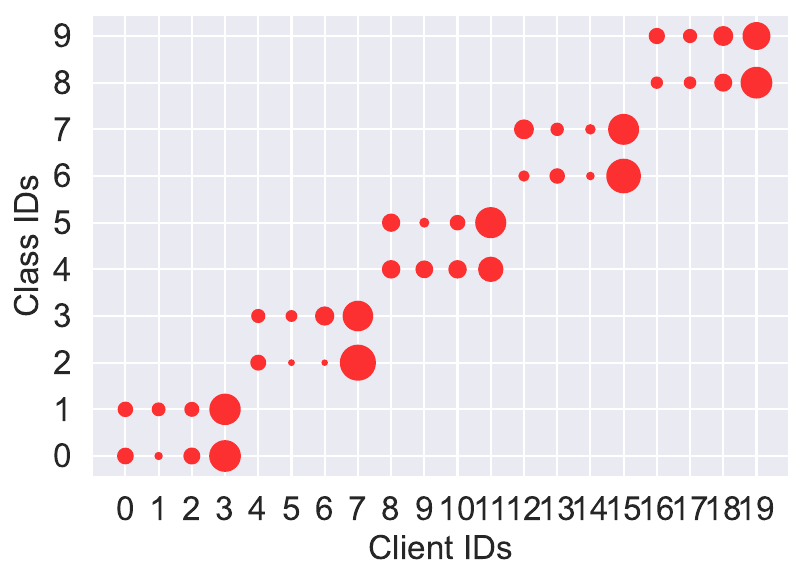}}
	\hfill
	\subfigure[Cifar10]{\includegraphics[width=0.32\linewidth]{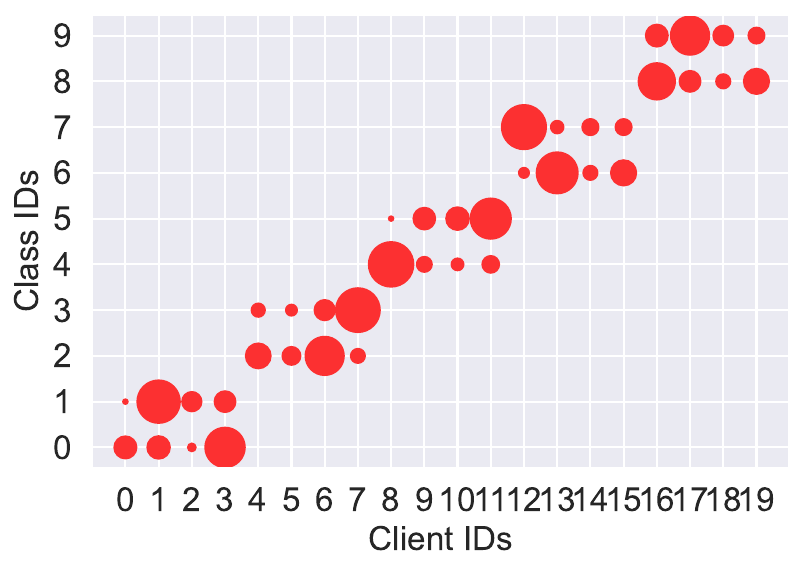}}
    \hfill
    \hfill
	\caption{The data distribution of each client on FMNIST and Cifar10, respectively, in the pathological settings. The size of a circle represents the number of samples. }
\end{figure*}

\begin{figure*}[ht]
	\centering
	\hfill
	\subfigure[FMNIST]{\includegraphics[width=0.32\linewidth]{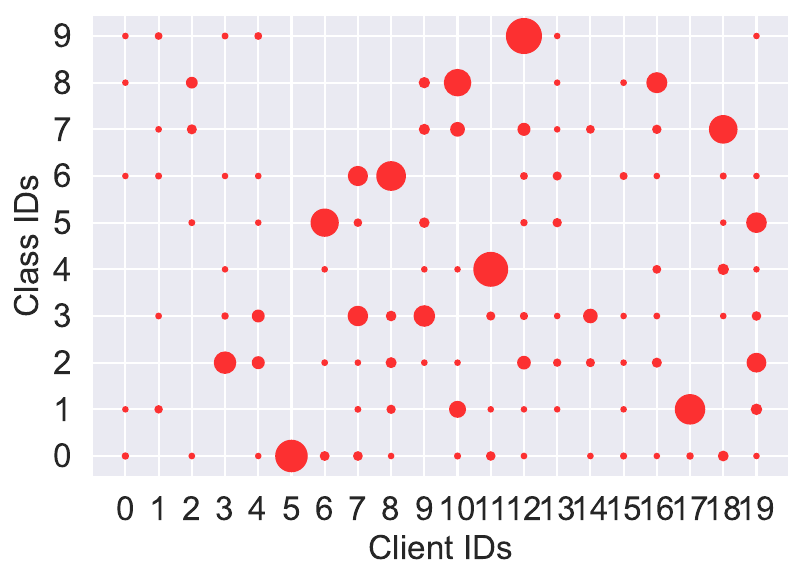}}
	\hfill
	\subfigure[Cifar10]{\includegraphics[width=0.32\linewidth]{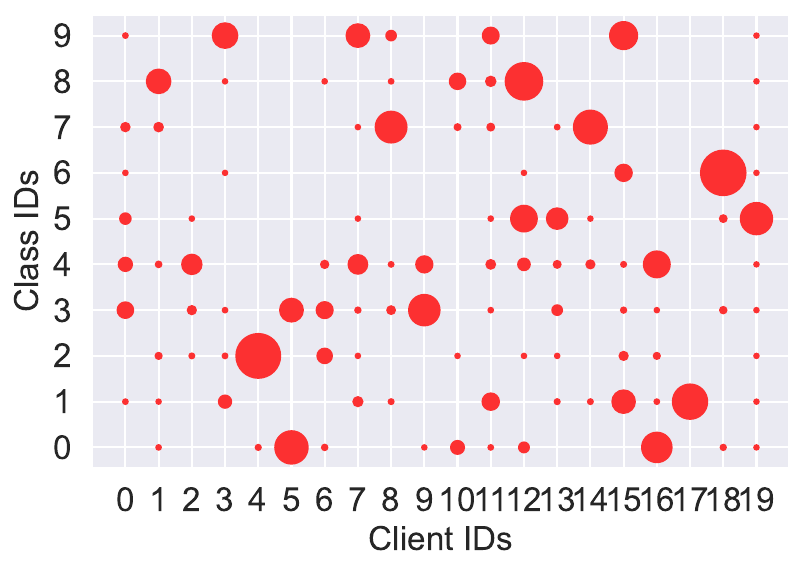}}
    \hfill
    \hfill
	\caption{The data distribution of each client on FMNIST and Cifar10, respectively, in practical settings ($\beta=0.1$). The size of a circle represents the number of samples. }
\end{figure*}

\begin{figure*}[ht]
	\centering
	\hfill
	\subfigure[Flowers102]{\includegraphics[width=0.32\linewidth]{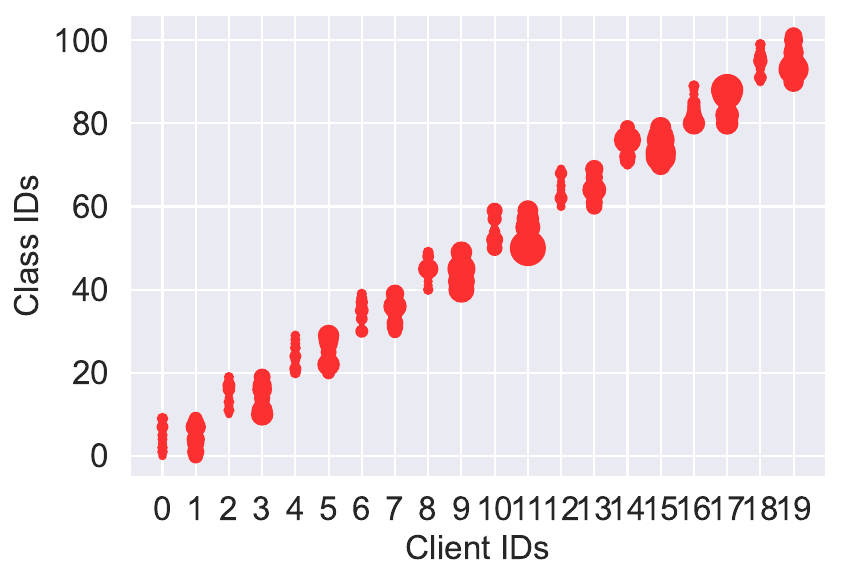}}
	\hfill
	\subfigure[Cifar100]{\includegraphics[width=0.32\linewidth]{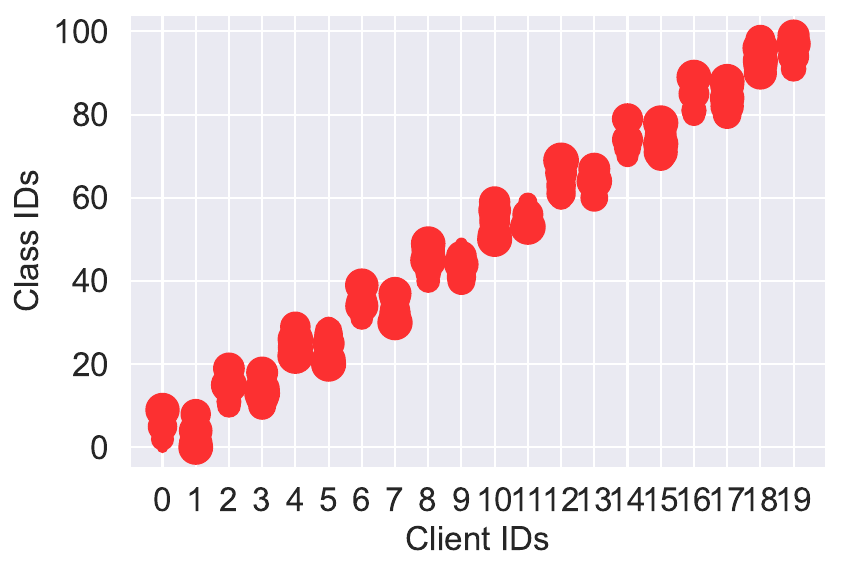}}
    \hfill
	\subfigure[Tiny-ImageNet]{\includegraphics[width=0.32\linewidth]{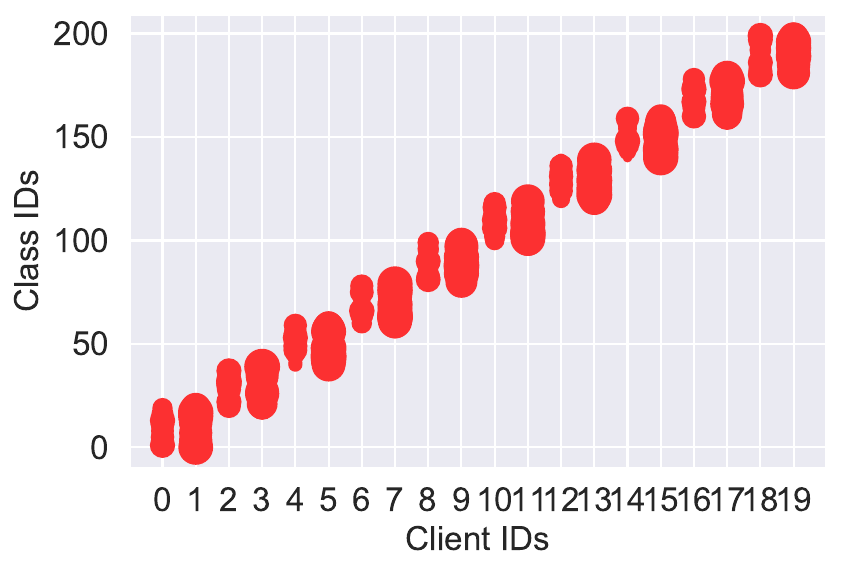}}
    \hfill
    \hfill
	\caption{The data distribution of each client on Flowers102, Cifar100, and Tiny-ImageNet, respectively, in the pathological settings. The size of a circle represents the number of samples. }
\end{figure*}

\begin{figure*}[ht]
	\centering
    \hfill
	\subfigure[Flowers102]{\includegraphics[width=0.32\linewidth]{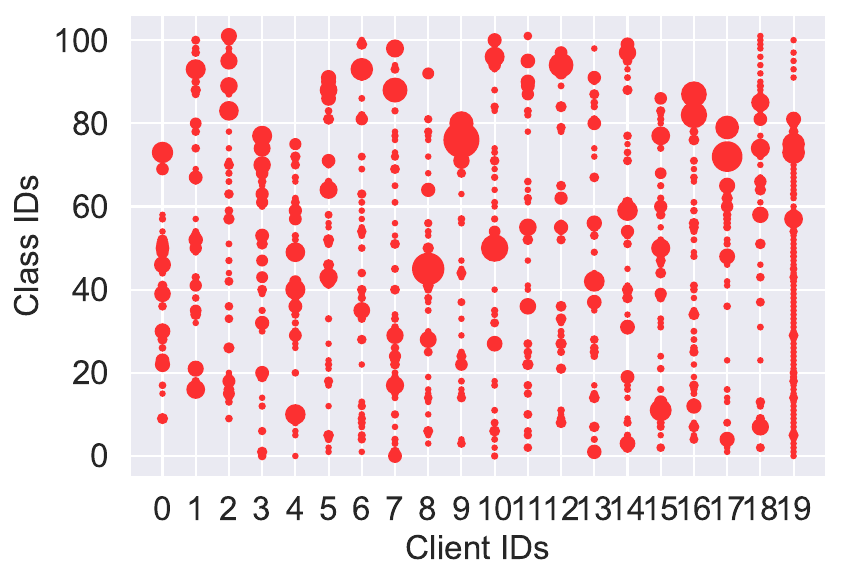}}
	\hfill
	\subfigure[Cifar100]{\includegraphics[width=0.32\linewidth]{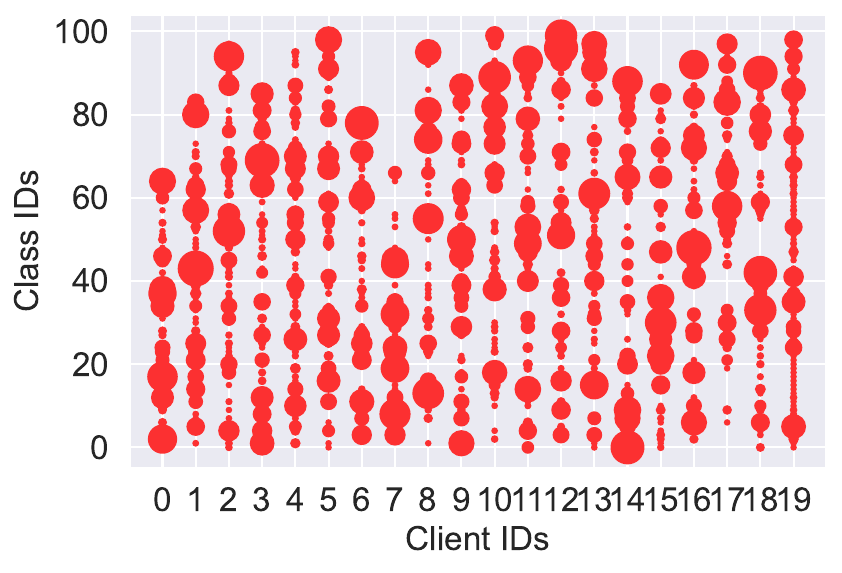}}
    \hfill
	\subfigure[Tiny-ImageNet]{\includegraphics[width=0.32\linewidth]{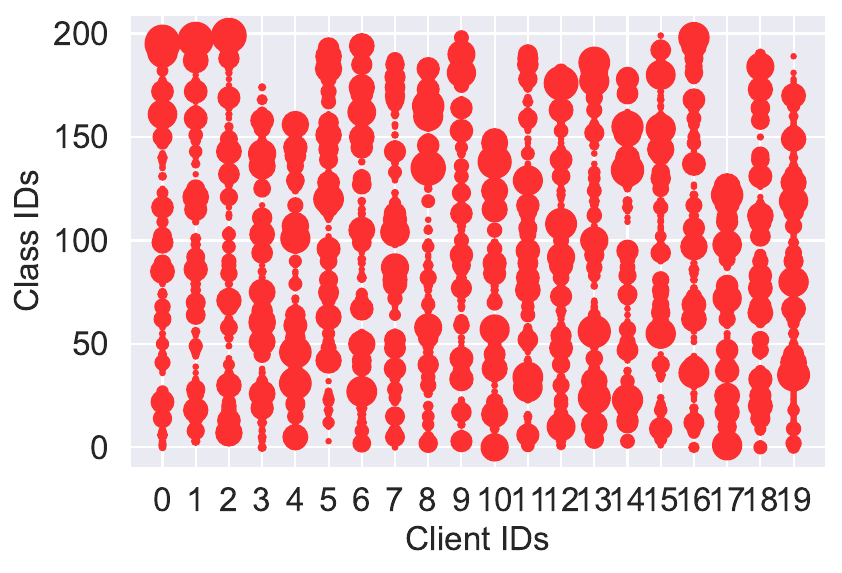}}
    \hfill
    \hfill
	\caption{The data distribution of each client on Flowers102, Cifar100, and Tiny-ImageNet, respectively, in practical settings ($\beta=0.1$). The size of a circle represents the number of samples. }
\end{figure*}

\begin{figure*}[ht]
	\centering
	\subfigure[50 clients]{\includegraphics[width=\linewidth]{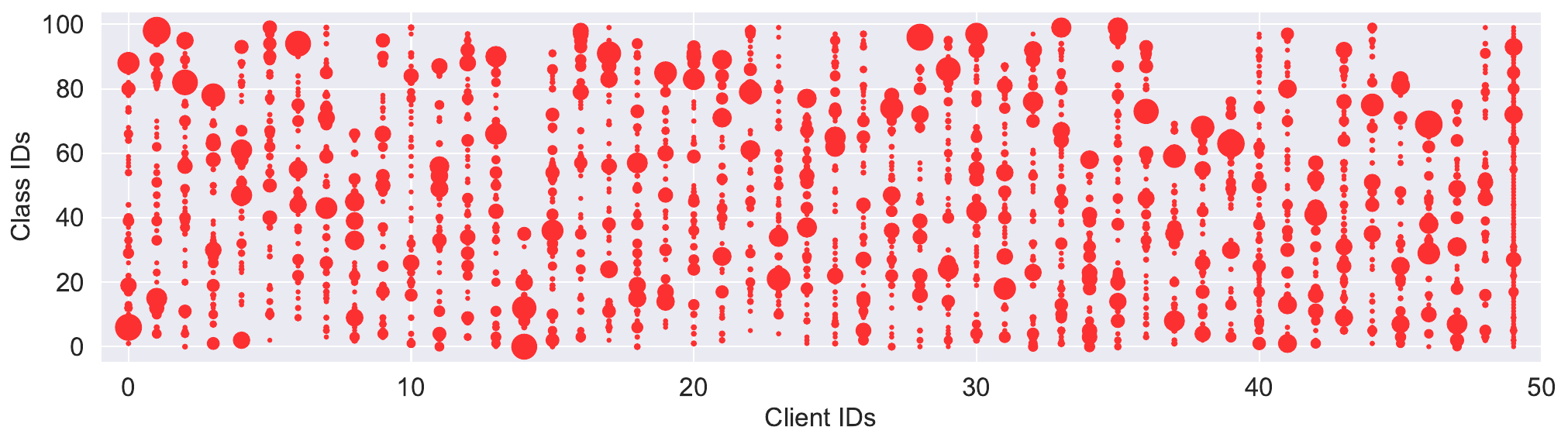}}
	\hfill
	\subfigure[100 clients]{\includegraphics[width=\linewidth]{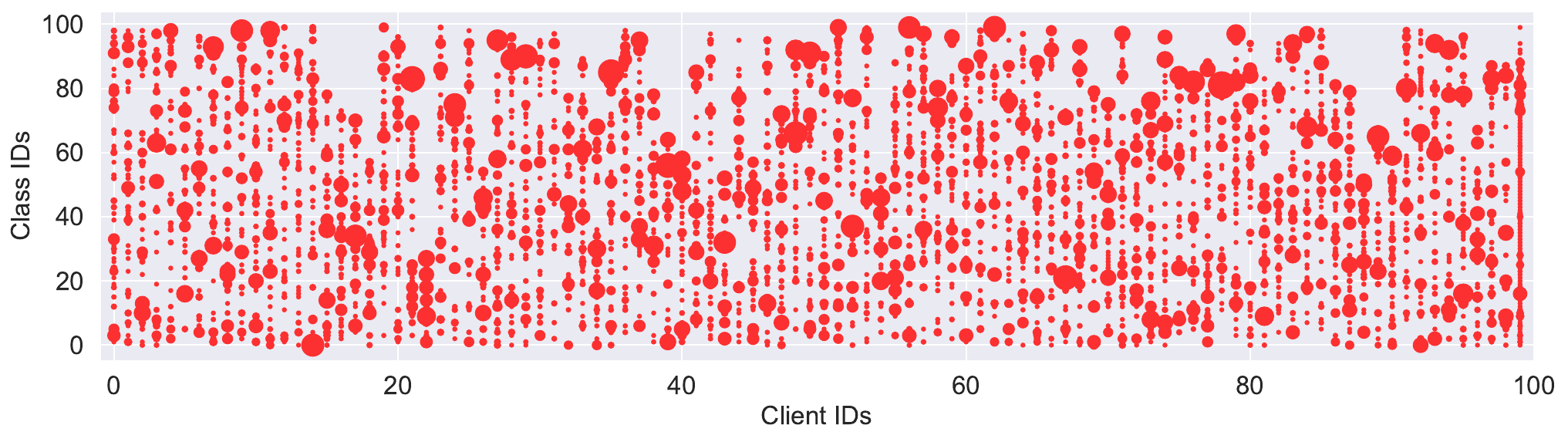}}
	\hfill
	\caption{The data distribution of each client on Cifar100 in the practical setting ($\beta=0.1$) with 50 and 100 clients, respectively. The size of a circle represents the number of samples. }
\end{figure*}

\end{document}